\documentclass[11pt]{article}
\usepackage[final]{acl}

\usepackage{times}
\usepackage{latexsym}
\usepackage[T1]{fontenc}
\usepackage[utf8]{inputenc}
\usepackage{inconsolata}
\usepackage{microtype}
\usepackage{graphicx}
\usepackage[most]{tcolorbox}
\usepackage{lipsum}
\usepackage{url}
\usepackage{hyperref}

\usepackage{amsmath}
\usepackage{amssymb}
\usepackage{booktabs}
\usepackage{multirow}
\usepackage{tabularx}
\usepackage{xspace}
\usepackage{enumitem}
\usepackage{array}

\usepackage{pgfplots} 
\pgfplotsset{compat=1.18}

\newcolumntype{Y}{>{\raggedright\arraybackslash}X}

\newcommand{\method}{CGM-Rec\xspace}

\newcommand{\Dtrain}{\mathcal{D}_{\mathrm{train}}}

\newcommand{\Dtest}{\mathcal{D}_{\mathrm{test}}}

\title{Continual Graph Memory for Adaptive Recommendation under \\ Intent Drift}

\author{
Hao Nguyen Ngoc\textsuperscript{1},
Tung Nguyen\textsuperscript{2,3},
Nguyen Thi Hanh\textsuperscript{1},
\\
\textbf{Hoang Thai Dinh}\textsuperscript{3},
\textbf{Nguyen Xuan Tung}\textsuperscript{1,$\dagger$}
\\
\textsuperscript{1}Phenikaa University, Vietnam,
\\
\textsuperscript{2}Hanoi University of Science and Technology, Vietnam,
\\
\textsuperscript{3}University of Technology Sydney, Australia
\\
\textsuperscript{$\dagger$}Corresponding author:{ tung.nguyenxuan@phenikaa-uni.edu.vn}
}

\begin{document}
\maketitle

\begin{abstract}
 \vspace{-0.3em}
This paper studies adaptive recommendation under intent drift, where feedback from each recommendation outcome can reveal whether the relational evidence used for ranking is useful, missing, or misleading. While Knowledge Graphs (KGs) provide essential semantic structure to handle these shifts, traditional KG-enhanced systems treat the graph as a static retrieval substrate, making it brittle to evolving intents, noisy metadata, and recurring failure patterns. This paper proposes \method\footnote{\url{https://anonymous.4open.science/r/CGM-17DD}}, a continual graph memory framework for adaptive recommendation. \method treats the graph state as a writable memory and maintains two complementary components. Therein, a Semantic Graph Memory is updated conservatively through quality-gated typed operations for storing stable and high-confidence relational knowledge. Meanwhile, an Episodic Lesson Memory acts as a fast reactive memory that learns recent outcomes, failure cases, and corrective hints. During testing, model parameters remain frozen and adaptation occurs only through memory writes. We evaluate \method under a frozen-parameter, one-pass reranking protocol, where encoders and prompts remain fixed during testing and adaptation occurs only through memory writes. Experiments across multiple recommendation settings show that \method improves over evaluated neural and LLM-based baselines on most metrics. Particularly, under sampled-candidate reranking, \method improves HR@1 by up to 29.58\% over the strongest LLM baseline on Bundle, and outperforms K-RagRec on metadata-rich ML-100K with HR@5 of 0.5941 versus 0.4746.
\end{abstract}
% These results suggest that adaptive recommendation benefits from maintaining graph evidence as long-term memory rather than only retrieving it.
% , suggesting that adaptive recommendation can benefit from maintaining graph evidence as long-term memory rather than only retrieving it.
 \vspace{-0.6em}
\section{Introduction}
 \vspace{-0.3em}
Recommendation systems are increasingly deployed in environments where user intents, item semantics, and feedback patterns change over time~\citep{hidasi2016gru4rec,tang2023dgel,sun2024po4isr}. In such settings, the main challenge is not only to infer preferences from past interactions, but also to decide which evidence should remain trusted as new outcomes arrive. Collaborative filtering alleviates information overload by learning latent user, item patterns, but it relies on dense interaction histories and remains vulnerable to sparsity~\cite{koren2009matrix,sarwar2001item}. Meanwhile, neural recommenders improve prediction by modeling high-dimensional, non-linear feature interactions~\cite{he2017neural,cheng2016wide}. Nevertheless, their knowledge is encoded implicitly in static parameters, limiting rapid adaptation to intent drift and providing little explicit structure for connecting sparse items. This raises a central question: \textit{how can a recommender preserve reusable knowledge while adapting to intent drift?}
% \vspace{-1em}
\begin{figure}[t]
    \centering
    \includegraphics[width=\columnwidth]{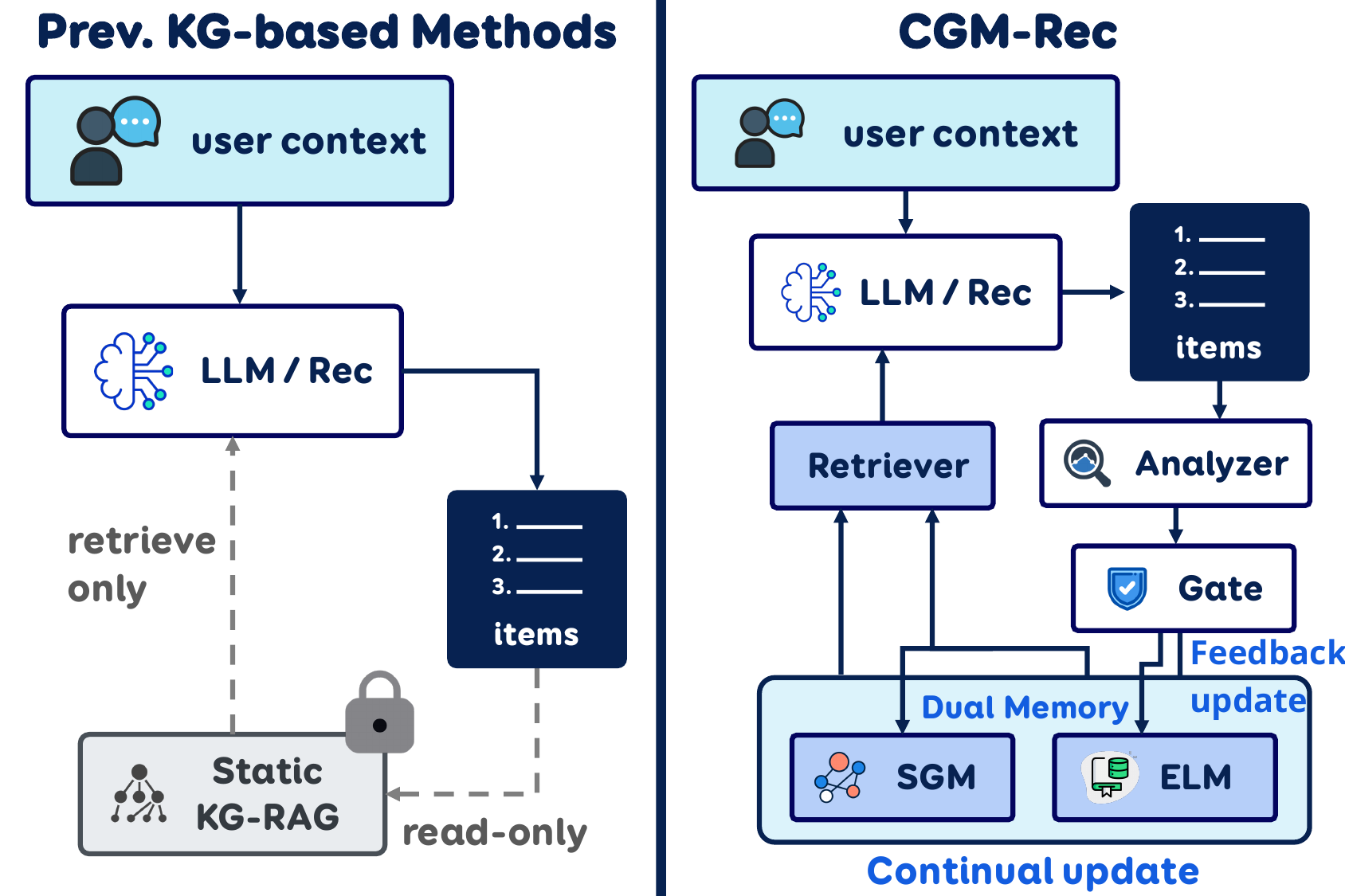}
    \vspace{-1.5em}
    \caption{Comparing read-only KG-based recommendation and the continual graph memory mechanism.}
    \vspace{-1.8em}
    \label{fig:intro_motivation}
\end{figure}
% \vspace{-1em}

Knowledge Graph (KG)-enhanced recommenders have emerged as a potential solution that incorporates external relational networks into the prediction process. By linking items to shared categories, descriptive keywords, and semantic entities, KGs provide a structured web of evidence that bridges sparse user-item interactions. Leveraging this structure, models like RippleNet~\cite{wang2018ripplenet} and KGAT~\cite{wang2019kgat} successfully enhance recommendation quality by propagating user preference signals along item-attribute paths. Graph-based session models, including SR-GNN~\citep{wu2019srgnn}, further show that graph reasoning can capture local transition patterns in short interaction sequences. However, traditional KG-based models are often constrained by predefined schemas and offline-learned embeddings, limiting their ability to interpret rich textual metadata and complex user intents. To bridge this gap, LLM-based recommenders have been widely adopted to improve semantic understanding and reasoning through prompting and in-context learning~\cite{wang2023nir,sun2024po4isr,li2023gpt4rec,yue2023llamarec}. More recently, KG-augmented LLM recommenders have exploited graph retrieval to improve semantic reasoning and ranking~\citep{chen2023kmvg,wang2025kragrec}. Despite these advances, most KG-enhanced recommenders treat the graph as a fixed retrieval source during inference. As illustrated in Figure~\ref{fig:intro_motivation}, this limits adaptive recommendation as read-only graphs keep retrieving noisy relations and cannot add or reinforce useful relations revealed by feedback. Therefore, under intent drift, the main question is: \textit{How to retrieve information from KG while maintaining accumulate knowledge?}

% This static approach creates a severe bottleneck for adaptive recommendation. Broad category relations can connect semantically distant items, sparse descriptions may render some item neighborhoods unreliable, and noisy metadata can introduce misleading graph paths. When a recommendation fails due to these misleading paths, a read-only graph cannot incorporate feedback, resulting in repeated errors as user intents change.

Existing adaptation methods address this issue incompletely. For instance, prompt optimization and LLM-based memory methods adapt instructions, reasoning traces, or retrieved text~\citep{wang2023nir,sun2024po4isr,nguyen2026llmgreenrec}. Meanwhile, continual recommendation and self-correction methods update model behavior, error compensation states, or learned parameters~\citep{cai2022reloop,zhu2023reloop2}. Dynamic graph methods model evolving relational structure~\citep{tang2023dgel}, but do not focus on frozen-parameter test-time adaptation through graph-memory writes. Consequently, the critical challenge remains: \textit{How to update relational evidence from feedback while preserving stable semantic knowledge and avoiding noisy or hallucinated graph edits?}

% Current adaptation methods address this inflexibility, but only indirectly. For instance, prompt optimization and error-memory techniques adapt the textual instructions provided to an LLM~\cite{sun2024po4isr,wang2023nir,nguyen2026llmgreenrec}, while continual recommendation approaches update model parameters~\cite{cai2022reloop,zhu2023reloop2}. Although these methods enhance adaptivity, they treat the underlying knowledge graph as a fixed substrate and adapt everything else around it. A critical challenge remains: how to directly maintain and evolve the knowledge graph itself as a long-term memory, enabling the system to correct noisy relations and consolidate new evidence without corrupting stable semantic knowledge.

Being inspired, the paper proposes \textbf{CGM-Rec} (\textbf{Continual Graph Memory for Recommendation}), a framework that treats the graph state as the primary adaptive object. Instead of retrieving from a fixed KG, CGM-Rec maintains a writable graph memory that can be updated after recommendation outcomes are observed. Specifically, the framework uses two complementary memories, including Semantic Graph Memory (SGM) and Episodic Lesson Memory (ELM). To swiftly capture emerging user intent drift, ELM acts as a fast-adapting buffer, recording recent recommendation feedback and correction hints through a quality-gated write policy acts as a strict filter. Meanwhile, SGM extracts and stores structural knowledge from data to ensure long-term robustness against noisy metadata and recurring failures. Results show that under a strict one-pass inference protocol with frozen parameters, CGM-Rec outperforms state-of-the-art baselines on most evaluated metrics. In summary, the key contributions of this paper are as follows:
\begin{itemize}[nosep, leftmargin=*]
    \item The paper formalizes adaptive recommendation under intent drift as an outcome-driven graph-memory maintenance problem, where recommendation feedback can be used to refine which relational evidence is useful, missing, or misleading for future ranking.
    \item We propose \method, a continual graph-memory framework that separates feedback by timescale and trust: ELM enables low-latency reuse of recent outcome-derived lessons, while SGM consolidates only supported, low-conflict evidence as persistent relational knowledge.

    \item We introduce a feedback-to-graph writing mechanism that converts recommendation outcomes into typed, provenance-aware graph-edit proposals and filters them using support, confidence, conflict, recency, and memory-cost signals before updating semantic memory.
    \item We evaluate \method under a frozen-parameter, one-pass sampled-candidate reranking protocol, showing improvements over evaluated neural and LLM-based baselines across multiple settings and over K-RagRec in the metadata-rich setting, with ablations supporting the complementary roles of episodic and semantic memory.
    % \item Introduces a dual-timescale memory architecture (ELM and SGM) to capture user intent drift while preserving stable structural knowledge.
    % \item We propose a continual graph-memory framework that makes the knowledge graph itself the adaptive state, separating fast feedback-derived lessons from conservative semantic consolidation to support adaptation without rewriting model parameters.

    % \item Develops a quality-gated write policy for outcome-aware graph edits, ensuring robust memory updates against noisy metadata.
    % \item Demonstrates CGM-Rec's effectiveness under one-pass inference, proving adaptation succeeds solely through continual memory maintenance.
\end{itemize}

% Conditionally promotes repeated low-conflict lessons from the ELM to the SGM, while actively pruning stale or misleading edges. This mechanism guarantees that the graph continuously purges noisy relations and adapts to the user's intents.

% While SGM stores stable, high-confidence relational knowledge, ELM stores recent outcomes, failure cases, and corrective signals. Feedback first enters episodic memory and can affect semantic memory only through a quality-gated write policy. This policy supports typed operations such as relation insertion, reinforcement, suppression, semantic promotion, and stale-edge pruning.

% By updating the graph memory from observed feedback through controlled operations, the system can dynamically prune misleading relations, reinforce successful paths, and immediately adapt to user intent drift without requiring model retraining.

% To achieve this effectively, CGM-Rec introduces a dual-timescale memory architecture that overcomes the aforementioned limitations. To swiftly capture emerging user intent drift, an Episodic Lesson Memory (ELM) acts as a fast-adapting buffer, recording recent recommendation feedback and correction hints. To ensure long-term robustness against noisy metadata and recurring failures, a Semantic Graph Memory (SGM) maintains stable, high-confidence structural knowledge.

\vspace{-0.3em}
\section{Related Work}
\vspace{-0.3em}
\subsection{Sequential and session-based recommendation.}
Session-based recommendation predicts the next item from a short interaction sequence, often without persistent user identity. Representative models include GRU4Rec~\citep{hidasi2016gru4rec}, NARM~\citep{li2017narm}, STAMP~\citep{liu2018stamp}, and SR-GNN~\citep{wu2019srgnn}. Sequential recommendation extends this setting to longer user histories, with Transformer-based models such as SASRec and BERT4Rec~\citep{kang2018sasrec,sun2019bert4rec}. These methods are effective at learning transition patterns from interaction sequences, but their adaptive state is mainly encoded in learned parameters or hidden sequence representations. As a result, they do not explicitly maintain external relational knowledge that can be corrected, consolidated, or pruned after recommendation outcomes.
% \vspace{-1em}
\subsection{KG-enhanced and KG-RAG recommendation.}
Knowledge-graph-enhanced recommenders use external relational structure to enrich item representations and connect sparse user-item interactions. For example, RippleNet~\citep{wang2018ripplenet} propagates user preferences over KG paths, while KGAT~\citep{wang2019kgat} performs relation-aware graph attention over collaborative knowledge graphs. Recent KG-augmented LLM recommenders, including K-RagRec~\citep{wang2025kragrec}, retrieve graph evidence to provide structured context for LLM-based generation or ranking. These methods demonstrate that graph knowledge can improve recommendation by grounding decisions in item-attribute and item-item relations. However, the graph is typically used as a retrieval source. Thus, recommendation outcomes may reveal missing, noisy, or misleading relations, but such feedback is not systematically written back into the graph as maintained memory. CGM-Rec differs by treating graph evidence as a writable memory state rather than only as retrieved context.
 % \vspace{-1em}
\subsection{LLM-based adaptive recommendation.}
LLMs have recently been used for recommendation through zero-shot ranking, prompt-based reasoning, and agent-style interaction. NIR~\citep{wang2023nir} studies zero-shot next-item recommendation, PO4ISR~\citep{sun2024po4isr} optimizes prompts for intent-driven session recommendation, and other LLM recommenders use generative ranking or multi-agent collaboration~\citep{li2023gpt4rec,yue2023llamarec,zhang2024agentcf,nguyen2026llmgreenrec}. Surveys and benchmarks further highlight the promise of LLM-based recommendation, while noting challenges in robustness, evaluation, memory design, and inference cost~\citep{peng2025survey,liu2025outshine}. These approaches improve semantic reasoning and can adapt through prompts, textual memories, or agent states. Nevertheless, their persistent adaptation is usually textual or prompt-level, not structured as typed graph memory. CGM-Rec instead uses the LLM as a frozen recommender and analyzer, while long-term adaptation is constrained to quality-gated graph-memory edits.
 % \vspace{-1em}
\subsection{Continual, dynamic, and memory-based recommendation.}
Dynamic memory-based recommendation methods aim to adapt recommenders as new interactions arrive. ReLoop \citep{cai2022reloop} and ReLoop2 \citep{zhu2023reloop2} introduce self-correction loops, while D2K \citep{qin2025d2k} turns historical data into a retrievable knowledge store. These works highlight the importance of continual adaptation and explicit memory. However, their adaptive states are error buffers, model updates, or generic retrieval stores rather than structured graph memories. Specifically, these methods do not explicitly model feedback as typed graph-edit operations to consolidate, weaken, or prune semantic relations over time. CGM-Rec is closest in spirit to this line, but its memory is organized as a graph with explicit operations for semantic promotion, relation reinforcement, suppression, and stale-edge pruning.

 \vspace{-0.3em}
\section{Task Definition}
 \vspace{-0.3em}
\label{sec:protocol}

Let $\mathcal{I}$ denote the item set and $\mathcal{A}$ denote available item attributes, such as categories, keywords, descriptions, and other metadata. We study adaptive recommendation as a one-pass candidate reranking problem over an ordered stream of instances $\mathcal{D}=\{(x_{t},\mathcal{C}_{t},y_{t})\}_{t=1}^{T}$, where $x_{t}$ is the context, $\mathcal{C}_{t}\subseteq\mathcal{I}$ denotes the candidate pool, and $y_{t}\in\mathcal{C}_{t}$ is the ground-truth target item. At each step $t$, the system maintains a dual-timescale memory state $\mathcal{M}_{t}=(\mathcal{M}_{s,t},\mathcal{M}_{e,t})$ , comprising a Semantic Graph Memory ($\mathcal{M}_{s,t}$) and an Episodic Lesson Memory ($\mathcal{M}_{e,t}$) (see Section \ref{sec:mem_cons} for details). The recommender leverages this state to generate a ranking list: $\hat{\pi}_{t}=f_{\theta}(x_{t},\mathcal{C}_{t},\mathcal{M}_{t})$. Once $\hat{\pi}_{t}$ is evaluated against $y_{t}$, the feedback is processed by a deterministic write policy to update the memory for the next step:
\begin{equation}
\mathcal{M}_{t+1} = U_{\phi}(\mathcal{M}_{t},x_{t},\mathcal{C}_{t},\hat{\pi}_{t},y_{t}),
\end{equation}
where fixed thresholds $\phi$ control typed operations. During inference, both recommender parameters $\theta$ and thresholds $\phi$ remain strictly frozen; adaptation is driven exclusively via memory updates.

\paragraph{Objective.}
The primary objective is to maximize stream-level ranking quality:
\begin{equation}
\mathcal{J} = \frac{1}{T}\sum_{t=1}^{T} \mathrm{Metric@K}(\hat{\pi}_{t}, y_{t}),
\end{equation}
where $\mathrm{Metric@K}$ denotes standard ranking indicators such as HR@K and NDCG@K (see Section \ref{sec:metrics}). While $\mathcal{J}$ isolates ranking effectiveness, the total memory capacity is governed by a strict, finite structural budget constraint rather than a continuous soft penalty function. When the memory footprint reaches its predefined capacity limit, the write policy $U_{\phi}$ automatically activates a deterministic cache eviction protocol. This mechanism systematically purges stale, low-utility, or highly conflicting historical records, maintaining bounded storage overhead and structural stability without sacrificing streaming recommendation accuracy.
 \vspace{-0.3em}
\section{Methodology}
 \vspace{-0.3em}

\begin{figure*}[ht]
  \centering
  \includegraphics[width=\textwidth]{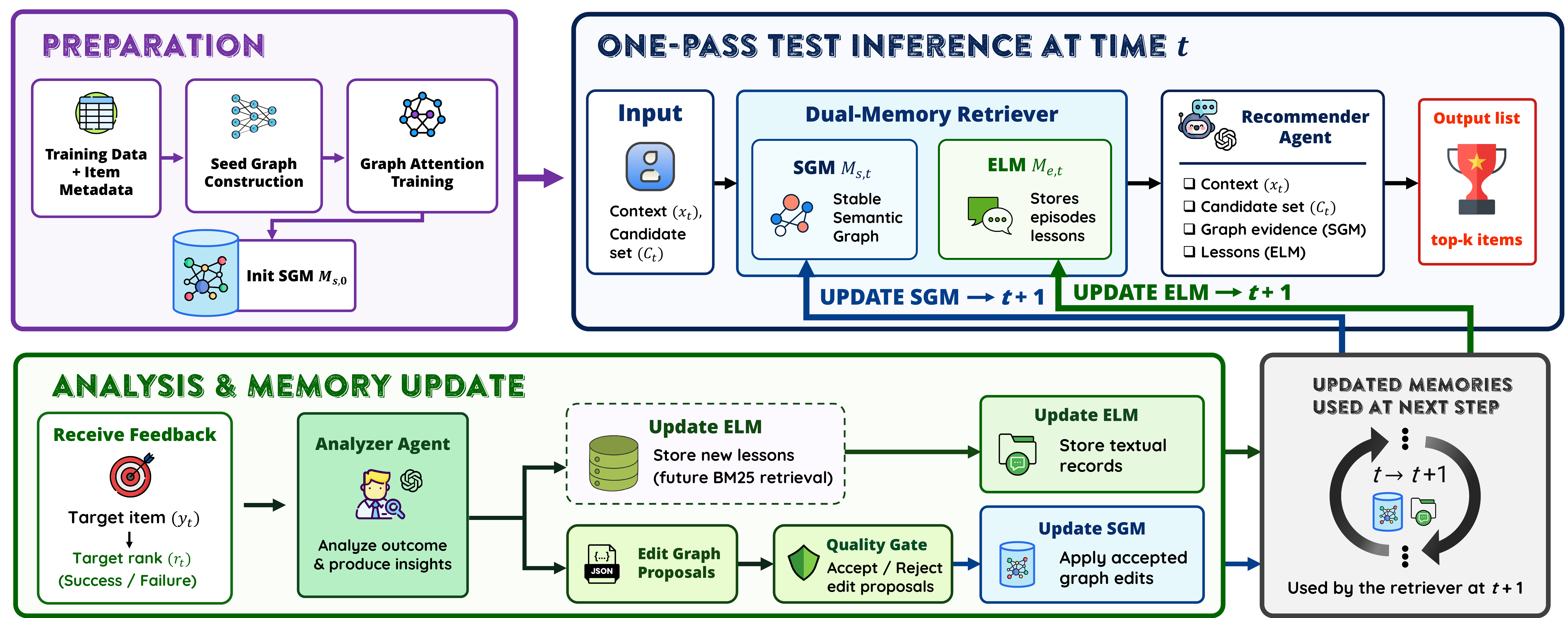}
 \vspace{-0.3em}
  \caption{Workflow of \method with continual dual-memory updates}
 \vspace{-0.6em}
  \label{fig:overall_workflow}
\end{figure*}
 % \vspace{-1em}
\subsection{Overview}

Figure~\ref{fig:overall_workflow} summarizes the workflow of \method. The framework has three stages. First, during offline preparation, training interactions and item metadata are used to construct a seed graph. A relation-aware graph encoder is trained on this seed graph and then frozen before test-time inference. The prompt templates and write-policy thresholds are fixed before testing. Second, at each test step \(t\), the system receives recommendation context \(x_t\) and candidate set \(\mathcal{C}_t\). A dual-memory retriever extracts semantic evidence from SGM and relevant textual lessons from ELM. These signals are combined into a structured prompt for a frozen LLM-based Recommender Agent, which reranks \(\mathcal{C}_t\) and outputs \(\hat{\pi}_t\). Third, after the target item \(y_t\) is revealed, the system analyzes the outcome. An Analyzer Agent produces textual lessons and structured graph-edit proposals. Textual lessons are stored in ELM for future retrieval, while graph-edit proposals must pass a fixed quality gate before modifying SGM. Thus, test-time adaptation is restricted to memory updates rather than parameter updates.

% Figure~\ref{fig:overall_workflow} illustrates the workflow of \method. Before inference, the system builds the initial graph memory from training interactions and item metadata, and fixes the graph encoder, prompt templates, and write thresholds. At each test step \(t\), \method receives a context \(x_t\) and candidate set \(C_t\). The dual-memory retriever reads subgraph evidence from Semantic Graph Memory (SGM) and retrieves top-\(k\) textual lessons from Episodic Lesson Memory (ELM) via BM25 \citep{robertson1994okapi}. The Recommender Agent then uses the context, candidates, graph evidence, and episodic lessons to rerank \(C_t\) and output a ranked list \(\hat{\pi}_t\). After the target \(y_t\) is revealed, an Analyzer Agent summarizes the outcome into textual lessons and structured graph-edit proposals. Lessons are stored in ELM for future retrieval, while edit proposals are filtered through a rule-guided quality gate before being updated in SGM. During the test stream, model parameters, prompt templates, and write thresholds remain frozen; adaptation is limited to memory contents: adding episodic lessons to ELM and applying accepted typed graph edits to SGM.
 % \vspace{-1em}
\subsection{Dual-Memory Architecture} \label{sec:mem_cons}

\method maintains two complementary memories. Semantic Graph Memory stores stable relational knowledge as a typed graph, while Episodic Lesson Memory stores recent outcome-derived lessons and correction signals. The two memories differ in trust level and update speed: ELM reacts quickly to recent feedback, whereas SGM changes conservatively through gated graph edits.

% \method constructs two complementary memories: a semantic graph memory for stable graph knowledge and an \textbf{Episodic Lesson Memory} (ELM) for textual lessons, outcome records, and structured signals that support future retrieval and graph updates. SGM provides relation-aware semantic evidence for recommendation, while ELM stores feedback-derived lessons that can be retrieved for similar future contexts.

\noindent \textbf{Seed graph construction.}
The system first samples a small subset of the training stream, together with item metadata, to construct a seed graph $G_0=(V_0,E_0,\mathcal{R}),$ where the node set $V_0$ contains item and metadata nodes (categories, keywords, descriptions). The typed edge set $E_0$ encodes relations over the vocabulary $\mathcal{R} = \{(i,\mathtt{belongs\_to},c),$ $(i,\mathtt{has\_keyword},k),$ $(i,\mathtt{has\_description},d),$ $(i,\mathtt{co\_occurs},j),$ and optionally $(i,\mathtt{related\_intent},z)\}$. The seed graph serves a dual purpose: initializing the Semantic Graph Memory ($\mathcal{M}_{s,0}$) and providing the structural substrate for a relation-aware graph encoder, which is trained using $D_{\mathrm{train}}$ and frozen before inference. At inference time, the frozen encoder is used only for graph-evidence retrieval, not for parameter updates. Details of the encoder and its training objective are provided in Appendix~\ref{app:graph_encoder}.

\noindent\textbf{Semantic Graph Memory.} \label{par:semantic_mem} SGM is initialized from the seed graph as $\mathcal{M}_{s,0}=(V_{s,0},E_{s,0},A_{s,0}),$ where \(V_{s,0}=V_0\), \(E_{s,0}=E_0\), and \(A_{s,0}\) stores edge attributes. Thus, at time \(t\), SGM is represented as $\mathcal{M}_{s,t}=(V_{s,t},E_{s,t},A_{s,t}).$ Each semantic edge \(e=(u,r,v)\in E_{s,t}\) is associated with:
\begin{equation}
A_{s,t}(e)=
(w_{e,t},c_{e,t},n_{e,t},q_{e,t},p_e),
\end{equation}
where \(w_{e,t}\) is the edge weight, \(c_{e,t}\) is confidence, \(n_{e,t}\) is the support count, \(q_{e,t}\) is a quality score, and \(p_e\) records provenance including metadata-derived, interaction-derived, or promoted-from-lesson. SGM stores relations that should change conservatively, including item-category, item-keyword, item-description, item-item, and accepted item-intent relations. During inference, SGM is updated only through accepted structured edit actions after the quality gate; raw LLM-generated text is never written directly into SGM.

\noindent \textbf{Episodic Lesson Memory.} \label{par:eps_mem}
ELM stores feedback-derived lessons and structured episodic signals generated after observing outcomes. Unlike the persistent, relation-typed SGM, ELM is a semi-structured store for recent outcome-specific evidence. At time \(t\), ELM denotes $\mathcal{M}_{e,t}=\{z_j\}_{j=1}^{B_t},$ where each record \(z_j\) contains lesson and structured metadata:
\begin{equation}
z_j=(d_j,\kappa_j,o_j,r_j,\mathcal{E}_j,\mathcal{P}_j,u_j,\rho_j).
\end{equation}
Here, $d_j$ is the lesson text, $\kappa_j$ the context, $o_j$ the outcome label, $r_j$ the target-rank signal, $\mathcal{E}_j$ the retrieved evidence, $\mathcal{P}_j$ the edit proposals, $u_j$ the utility or support, and $\rho_j$ the retrieval metadata. This design separates short-term episodic adaptation from long-term semantic consolidation. ELM preserves recent success patterns, failure causes, misleading evidence, and corrective hints for future retrieval, but its records do not directly modify SGM. Only structured proposals accepted by the quality-gated write policy can update SGM.
 % \vspace{-1em}
\subsection{Dual-memory recommendation.} 
Given context \(x_t\) and candidate set \(C_t\), the system retrieves complementary evidence from SGM and ELM before reranking. For SGM, context and candidate items are mapped to graph nodes. The retriever then expands local neighborhoods around these nodes in \(\mathcal{M}_{s,t}\) and uses the frozen graph encoder to rank local edges and relational paths based on their relevance to \((x_t,C_t)\). This produces:
\begin{equation}
    R_t^s=\mathrm{GraphRetrieve}(x_t,C_t,\mathcal{M}_{s,t};\theta_g),
\end{equation}
where \(\theta_g\) denotes the frozen graph encoder. The retrieved subgraph evidence includes item--attribute relations, item--item links, intent-related relations, and edge attributes like confidence, support, and provenance. For ELM retrieval, the system constructs a query from the current context, candidate items, and available item attributes, \(q_t=\mathrm{Query}(x_t,C_t)\), and retrieves top-\(k\) relevant lessons using BM25: $R_t^e = \mathrm{TopK}_{\mathcal{M}_{e,t}} \mathrm{BM25}(q_t,d_j,\kappa_j)$.

These complementary signals are directly integrated to populate a structured context prompt $P_t = \mathrm{Prompt}(x_t,\mathcal{C}_t,R_t^s,R_t^e)$, fusing stable relational structures with experiences (see Appendix~\ref{app:rec_temp} for full prompt templates). This prompt is processed by the Recommender Agent, an LLM-based reranker governed by a specific task instruction. The agent is instructed to analyze the user context, infer the current intent, and leverage the retrieved SGM and ELM evidence to evaluate the candidates. Operating under strictly frozen weights without test-time fine-tuning or gradient updates, the agent outputs the optimized ranking $\hat{\pi}_t \in \Pi(\mathcal{C}_t)$:
\begin{equation}
\hat{\pi}_t = \mathrm{Rerank}_{\omega}(\mathcal{C}_t \mid x_t, R_t^s, R_t^e),
\end{equation}
where $\omega$ denotes the fixed LLM prompt configuration, and $\Pi(\mathcal{C}_t)$ denotes all candidate permutations.
 % \vspace{-1em}
\subsection{Quality-Gated Memory Update}

After the recommender outputs \(\hat{\pi}_t\), the target item \(y_t\) becomes available for evaluation and memory update. Importantly, \(y_t\) is not used before the ranking is produced. The system computes the target rank $r_t=\mathrm{rank}_{\hat{\pi}_t}(y_t)$ and assigns a binary outcome label $o_t \in \{\texttt{success}, \texttt{failure}\}$ against a predefined rank cutoff threshold $K_s$, where $o_t = \texttt{success}$ if $r_t \le K_s$ and $\texttt{failure}$ otherwise. This label provides a compact feedback signal for subsequent lesson generation and memory updates.

\noindent \textbf{Outcome analysis.}
After obtaining the target rank and outcome label, the system packages the context, retrieved evidence, and feedback into:
$$\Omega_t = (x_t,\mathcal{C}_t,\hat{\pi}_t,y_t,r_t,o_t,R_t^s,R_t^e)$$ 
An LLM-based Analyzer Agent (prompt in Appendix~\ref{app:anal_temp}) processes this package: $(\mathcal{L}_t,\mathcal{P}_t) = \mathrm{Analyzer}_{\omega_a}(\Omega_t)$,
where \(\mathcal{L}_t\) denotes textual lessons and \(\mathcal{P}_t\) denotes structured graph-edit proposals. The Analyzer is instructed to explain the observed outcome, identify useful or misleading evidence, and propose candidate memory edits in a constrained format; it cannot directly modify SGM. The lessons summarize intent cues, useful patterns, failure causes, misleading evidence, or corrective hints and are stored in ELM for future retrieval:
\begin{equation}
\mathcal{M}_{e,t+1}
=
\mathrm{Budget}(\mathcal{M}_{e,t}\cup \mathcal{L}_t),
\end{equation}
The \(\mathrm{Budget}(\cdot)\) operator keeps ELM within a fixed capacity by retaining recent and useful lessons while removing low-utility records.

\noindent \textbf{Structured edit proposals.}
Each proposal \(p\in\mathcal{P}_t\) is represented as $p=(a,z,\delta,\xi),$ where \(a\) is the action type, \(z\) is the target object such as an edge, relation, or lesson-derived pattern, \(\delta\) stores the update value or arguments, and \(\xi\) stores supporting evidence from the outcome package. The supported action types are summarized in Table~\ref{tab:edits}. These proposals are treated as candidates only and can affect SGM only after passing the quality gate.

\begin{table*}[t]
\centering
\small
\begin{tabularx}{\linewidth}{llX}
\toprule
\textbf{Action} & \textbf{Target} & \textbf{Function} \\
\midrule
\texttt{add\_tentative\_edge} 
& SGM 
& Add a low-confidence relation when feedback suggests missing evidence. \\

\texttt{reinforce\_edge} 
& SGM 
& Increase the weight, confidence, or support count of a relation that improves ranking. \\

\texttt{suppress\_edge} 
& SGM 
& Downweight a relation that misleads ranking or conflicts with observed feedback. \\

\texttt{store\_lesson} 
& ELM 
& Store a textual lesson or corrective hint generated from the outcome analysis. \\

\texttt{promote\_to\_semantic} 
& SGM 
& Convert a reliable recurring lesson or edit proposal into a stable semantic relation. \\

\texttt{prune\_stale\_edge} 
& SGM 
& Remove or deactivate a low-confidence, low-utility, or stale relation. \\
\bottomrule
\end{tabularx}
 \vspace{-0.5em}
\caption{Typed actions used by the feedback-guided memory update mechanism.}
 \vspace{-1.6em}
\label{tab:edits}
\end{table*}

\noindent\textbf{Quality-gated writing.}
To prevent unsupported LLM proposals from modifying SGM, structured proposals are filtered by a rule-guided quality gate \(U_{\phi}\). The gate is a deterministic write filter with predefined thresholds \(\phi\), rather than a learned model. For each proposal \(p\), it checks rank, support, confidence, SGM conflict, memory cost, and recency:
\begin{equation}
\label{eq:write_signals}
\begin{aligned}
\mathcal{S}_t(p)=\{
&o_t, r_t,
\mathrm{supp}(p), \mathrm{conf}(p), \\ &
\mathrm{conflict}(p), \mathrm{cost}(p),
\mathrm{rec}(p)
\}.
\end{aligned}
\end{equation}
The quality gate filters proposals into the accepted action set $\mathcal{A}_t=U_{\phi}(\mathcal{P}_t,\mathcal{S}_t,\mathcal{M}_{s,t})$ using thresholds fixed before test-time inference, reinforcing useful relations while suppressing misleading ones (see Appendix~\ref{app:quality_gate} for detailed criteria).

\noindent \textbf{Updating SGM.}
Only accepted typed actions are applied to Semantic Graph Memory:
\begin{equation}
\mathcal{M}_{s,t+1}
=
\mathrm{Apply}(\mathcal{M}_{s,t},\mathcal{A}_t).    
\end{equation}
The \(\mathrm{Apply}(\cdot)\) operator executes the accepted action types in Table~\ref{tab:edits}. Reinforcement increases the support, confidence, or weight of useful relations, while suppression decreases the confidence or weight of relations associated with repeated failures or misleading evidence. Tentative edges are inserted with low confidence and must accumulate support to become stable relations. Finally, weak, unsupported, or low-utility edges are pruned to keep SGM compact and prevent noise accumulation over long streams (details in Appendix~\ref{app:sgm_update}).
% \vspace{-0.6em}
\section{Experiments}
% \vspace{-0.3em}

The experiments address four research questions: \textbf{(RQ1)} Does \method outperform traditional, neural, and LLM-based baselines? \textbf{(RQ2)} How does \method compare to LLM-based methods in metadata-rich contexts? (\textbf{RQ3}) How do dual memory and graph updating affect performance? (\textbf{RQ4}) How does \method adapt under continual deployment and measurable intent shifts?
 % \vspace{-0.3em}
\subsection{Experimental Setup}
\noindent \textbf{Dataset.} \method is evaluated on four datasets: Bundle~\citep{Zhu2022BundleDataset}, Games~\citep{Ni2019Justifying}, and MovieLens ML-1M/ML-100K~\citep{Harper2015movielens}. Bundle and Games represent short-context product and game recommendation settings, ML-1M provides user-movie interaction contexts, and ML-100K serves as a metadata-rich non-session contextual reranking benchmark with taxonomy paths, descriptions, and keywords. Each instance contains a context, a candidate set, and a ground-truth target item. Each context is an interaction sequence or a metadata-rich textual query, and the candidate set includes the ground-truth target item along with sampled items for reranking. Therefore, our evaluation focuses on constrained candidate reranking rather than full-catalog retrieval. All methods rerank the same candidate sets, and Table~\ref{tab:data_stats} reports dataset statistics.

\noindent \textbf{Evaluation Protocol.} Each dataset is split into training \(\Dtrain\) and test stream \(\Dtest\). A small subset of \(\Dtrain\), together with item metadata, is used to construct the seed graph and initialize \(\mathcal{M}_{s,0}\), while \(\Dtrain\) is used for offline graph-encoder training. This reflects a realistic deployment setting where the system starts with sparse initial knowledge and gradually accumulates structural memory from new recommendation outcomes. All write-policy thresholds are fixed before testing and are not tuned on \(\Dtest\). During one-pass inference, model parameters and prompt templates remain frozen. Each instance follows:
\vspace{-0.2em}
\begin{equation}
\mathrm{retrieve}\rightarrow\mathrm{rank}\rightarrow\mathrm{feedback}\rightarrow\mathrm{update}.
\end{equation}
\vspace{-0.2em}
The target item is revealed after ranking to compute feedback and update memory for future instances. All main comparisons are repeated over five matched seeds $\{0, 10, 42, 625, 2023\}$, and the results are reported as mean $\pm$ standard deviation.

\noindent \textbf{Leakage control.} All continual evaluations follow a strict predict--reveal--update protocol. At step $t$, the recommendation is produced using the context and memory state available before observing the target outcome. The target and its feedback are revealed only after ranking, and any SGM or ELM update can affect subsequent steps. To prevent target leakage through episodic memory, stored lessons are sanitized to exclude target IDs and titles, candidate positions, exact ranks, raw Analyzer outputs, and executable update proposals; only attribute- and relation-level summaries are retained.

\noindent \textbf{Inference overhead.} CGM-Rec averages 9.6\,s per recommendation instance, including both LLM calls and memory operations, with an average of 19.93K input and 1.17K output tokens. (breakdown in Appendix~\ref{sec:appendix_efficiency}).

\noindent \textbf{Metrics.} \label{sec:metrics}
Two ranking metrics are used:  Hit
Rate (HR@K) and Normalized Discounted Cumulative Gain
(NDCG@K), with \(K \in \{1,5,10\}\)~\citep{sun2024po4isr}. HR@K measures whether the ground-truth item appears within the top-\(K\) positions, while NDCG@K further accounts for its ranking position. Higher values indicate better performance. 
% For each instance \(t\), the metrics are defined as:
% \begin{equation}
% \begin{aligned}
% \mathrm{HR@K}_t
% &= \mathbb{I}\left[\mathrm{rank}_{\hat{\mathbf{r}}_t}(y_t) \le K\right], \\
% \mathrm{NDCG@K}_t
% &= \frac{\mathbb{I}\left[\mathrm{rank}_{\hat{\mathbf{r}}_t}(y_t) \le K\right]}
% {\log_2\left(\mathrm{rank}_{\hat{\mathbf{r}}_t}(y_t)+1\right)} .
% \end{aligned}
% \end{equation}

\subsection{Baselines}

\method is benchmarked against a wide range of baselines spanning four distinct paradigms. The first category encompasses traditional and deep learning-based single-intent recommenders: MostPop~\cite{ji2020popularityBaseline} for popularity-based ranking, SKNN~\cite{diet2017SKNN} for neighborhood similarity, FPMC~\cite{stef2010FPMC} for Markov chain modeling, and attention-based neural models (NARM~\cite{li2017narm}, STAMP~\cite{liu2018stamp}). This category also includes SASRec~\cite{kang2018sasrec}, a self-attentive sequential recommender, and GCE-GNN~\cite{ziyang2023GCE-GNN}, a session recommender employing graph neural networks to aggregate item transitions across both local sessions and a global context graph. The second category comprises advanced multi-intent frameworks, including MCPRN~\cite{wang2019MCPRN}, HIDE~\cite{li2022HIDE}, and Atten-Mixer~\cite{pei2022Atten-mixer}. These models explicitly route or mix sequence representations into multiple intention channels to handle diverse behavioral patterns. For continual self-adaptation, ReLoop2~\cite{zhu2023reloop2} is included, which adapts online through a responsive error-compensation loop without LLM components. The final category includes recent LLM-based baselines: NIR~\cite{wang2023nir}, PO4ISR~\cite{sun2024po4isr}, LLMGreenRec~\cite{nguyen2026llmgreenrec}, and K-RagRec~\citep{wang2025kragrec}, covering zero-shot reasoning, prompt optimization, textual memory retrieval, and graph-augmented LLM recommendation. To ensure fairness, all LLM-based methods use GPT-4.1-mini as the backbone and are evaluated on the same candidate set of size 20.
 \vspace{-0.3em}
\section{Experiment Results}
 \vspace{-0.3em}
\subsection{Overall Performance Comparison}

\begin{table*}[t]
  \centering
  \scriptsize
  \setlength{\tabcolsep}{1.6pt}
  \renewcommand{\arraystretch}{1.2}
  \resizebox{\textwidth}{!}{%
  \begin{tabular}{l l ccc cccc ccc ccccc}
    \toprule
    \multirow{2}{*}{\textbf{Datasets}} & \multirow{2}{*}{\textbf{Metrics}}
      & \multicolumn{3}{c}{\textbf{Traditional}}
      & \multicolumn{4}{c}{\textbf{Single-Intent}}
      & \multicolumn{3}{c}{\textbf{Multi-Intent}}
      & \multicolumn{5}{c}{\textbf{Continual \& LLM-based}} \\
    \cmidrule(lr){3-5}\cmidrule(lr){6-9}\cmidrule(lr){10-12}\cmidrule(lr){13-17}
    & & MostPop & SKNN & FPMC
      & NARM & STAMP & GCE-GNN & SASRec
      & MCPRN & HIDE & Atten-Mixer
      & ReLoop2 & NIR & PO4ISR & LLMGreenRec & \textbf{\method} \\
    \midrule
    \multirow{3}{*}{ML-1M}
      & HR@1
        & 0.0071{\tiny$\pm$.0016} & 0.1502{\tiny$\pm$.0083} & 0.1254{\tiny$\pm$.0109}
        & 0.0670{\tiny$\pm$.0046} & 0.1702{\tiny$\pm$.0101} & 0.1314{\tiny$\pm$.0046} & \underline{0.2018}{\tiny$\pm$.0136}
        & 0.0794{\tiny$\pm$.0062} & 0.1498{\tiny$\pm$.0090} & 0.1380{\tiny$\pm$.0087}
        & 0.1132{\tiny$\pm$.0076} & 0.0564{\tiny$\pm$.0348} & 0.1786{\tiny$\pm$.0273} & 0.1633{\tiny$\pm$.0389} & \textbf{0.2049}{\tiny$\pm$.0117} \\
      & HR@5
        & 0.0527{\tiny$\pm$.0052} & 0.3376{\tiny$\pm$.0121} & 0.3804{\tiny$\pm$.0040}
        & 0.3054{\tiny$\pm$.0066} & 0.5278{\tiny$\pm$.0111} & 0.4818{\tiny$\pm$.0069} & 0.5618{\tiny$\pm$.0140}
        & 0.3374{\tiny$\pm$.0094} & 0.5096{\tiny$\pm$.0064} & 0.4986{\tiny$\pm$.0075}
        & 0.3532{\tiny$\pm$.0154} & 0.2314{\tiny$\pm$.0994} & \underline{0.5374}{\tiny$\pm$.0426} & 0.5111{\tiny$\pm$.0655} & \textbf{0.5786}{\tiny$\pm$.0174} \\
      & NDCG@5
        & 0.0353{\tiny$\pm$.0053} & 0.2429{\tiny$\pm$.0098} & 0.2554{\tiny$\pm$.0052}
        & 0.1822{\tiny$\pm$.0056} & 0.3517{\tiny$\pm$.0062} & 0.3057{\tiny$\pm$.0054} & \underline{0.3692}{\tiny$\pm$.0110}
        & 0.2085{\tiny$\pm$.0055} & 0.3287{\tiny$\pm$.0069} & 0.3189{\tiny$\pm$.0051}
        & 0.2338{\tiny$\pm$.0088} & 0.1445{\tiny$\pm$.0480} & 0.3601{\tiny$\pm$.0412} & 0.3537{\tiny$\pm$.0286} & \textbf{0.3878}{\tiny$\pm$.0156} \\
    \midrule
    \multirow{3}{*}{Games}
      & HR@1
        & 0.0058{\tiny$\pm$.0037} & 0.0498{\tiny$\pm$.0078} & 0.0602{\tiny$\pm$.0054}
        & 0.0584{\tiny$\pm$.0038} & 0.0622{\tiny$\pm$.0055} & 0.0590{\tiny$\pm$.0016} & 0.0988{\tiny$\pm$.0044}
        & 0.0608{\tiny$\pm$.0044} & 0.0780{\tiny$\pm$.0029} & 0.0594{\tiny$\pm$.0011}
        & 0.2336{\tiny$\pm$.0126} & 0.1055{\tiny$\pm$.0463} & 0.2602{\tiny$\pm$.0561} & \underline{0.2739}{\tiny$\pm$.0402} & \textbf{0.3092}{\tiny$\pm$.0131} \\
      & HR@5
        & 0.0259{\tiny$\pm$.0042} & 0.2534{\tiny$\pm$.0084} & 0.2544{\tiny$\pm$.0062}
        & 0.2644{\tiny$\pm$.0092} & 0.2672{\tiny$\pm$.0018} & 0.2656{\tiny$\pm$.0021} & 0.2920{\tiny$\pm$.0117}
        & 0.2574{\tiny$\pm$.0058} & 0.2674{\tiny$\pm$.0078} & 0.2578{\tiny$\pm$.0069}
        & \textbf{0.5962}{\tiny$\pm$.0177} & 0.4310{\tiny$\pm$.0704} & 0.5474{\tiny$\pm$.0616} & 0.5494{\tiny$\pm$.0385} & \underline{0.5898}{\tiny$\pm$.0176} \\
      & NDCG@5
        & 0.0155{\tiny$\pm$.0043} & 0.1481{\tiny$\pm$.0070} & 0.1550{\tiny$\pm$.0034}
        & 0.1568{\tiny$\pm$.0059} & 0.1634{\tiny$\pm$.0020} & 0.1592{\tiny$\pm$.0015} & 0.1943{\tiny$\pm$.0058}
        & 0.1568{\tiny$\pm$.0034} & 0.1701{\tiny$\pm$.0052} & 0.1558{\tiny$\pm$.0029}
        & 0.4194{\tiny$\pm$.0151} & 0.2656{\tiny$\pm$.0462} & 0.4153{\tiny$\pm$.0576} & \underline{0.4236}{\tiny$\pm$.0340} & \textbf{0.4475}{\tiny$\pm$.0168} \\
    \midrule
    \multirow{3}{*}{Bundle}
      & HR@1
        & 0.0038{\tiny$\pm$.0059} & 0.0546{\tiny$\pm$.0066} & 0.0261{\tiny$\pm$.0035}
        & 0.0538{\tiny$\pm$.0100} & 0.0345{\tiny$\pm$.0120} & 0.0571{\tiny$\pm$.0048} & 0.0261{\tiny$\pm$.0075}
        & 0.0311{\tiny$\pm$.0105} & 0.0538{\tiny$\pm$.0117} & 0.0630{\tiny$\pm$.0079}
        & 0.1454{\tiny$\pm$.0114} & 0.0980{\tiny$\pm$.0133} & 0.1657{\tiny$\pm$.0099} & \underline{0.2518}{\tiny$\pm$.0229} & \textbf{0.3263}{\tiny$\pm$.0175} \\
      & HR@5
        & 0.0227{\tiny$\pm$.0021} & 0.2605{\tiny$\pm$.0195} & 0.2261{\tiny$\pm$.0196}
        & 0.2630{\tiny$\pm$.0189} & 0.2487{\tiny$\pm$.0237} & 0.2723{\tiny$\pm$.0035} & 0.2185{\tiny$\pm$.0390}
        & 0.2025{\tiny$\pm$.0137} & 0.2672{\tiny$\pm$.0097} & 0.2832{\tiny$\pm$.0064}
        & 0.4185{\tiny$\pm$.0304} & 0.2882{\tiny$\pm$.0901} & 0.3908{\tiny$\pm$.0467} & \underline{0.5160}{\tiny$\pm$.0086} & \textbf{0.5716}{\tiny$\pm$.0098} \\
      & NDCG@5
        & 0.0125{\tiny$\pm$.0050} & 0.1535{\tiny$\pm$.0091} & 0.1199{\tiny$\pm$.0082}
        & 0.1567{\tiny$\pm$.0117} & 0.1347{\tiny$\pm$.0053} & 0.1583{\tiny$\pm$.0043} & 0.1174{\tiny$\pm$.0246}
        & 0.1133{\tiny$\pm$.0077} & 0.1600{\tiny$\pm$.0106} & 0.1677{\tiny$\pm$.0075}
        & 0.2831{\tiny$\pm$.0198} & 0.1974{\tiny$\pm$.0515} & 0.2886{\tiny$\pm$.0379} & \underline{0.3996}{\tiny$\pm$.0123} & \textbf{0.4495}{\tiny$\pm$.0146} \\
    \bottomrule
  \end{tabular}}
  \caption{Performance comparison on three datasets, reported as mean\,$\pm$\,SD over 5 seeds. Best in bold, runner-up underlined. Paired $t$-tests against the runner-up give $p=$ 7.07e-03 (Bundle HR@1), 1.99e-03 (Bundle HR@5); differences on Games and ML-1M are not statistically significant ($p>0.1$).}
  \label{tab:overall_comparison}
\end{table*}

Table~\ref{tab:overall_comparison} shows that \method achieves the best results on most datasets and metrics, demonstrating strong ranking performance across the evaluated settings. Compared to traditional and neural sequence recommenders, \method yields substantial gains because static paradigms fail to maintain adaptive states during inference. This limitation is clear on the Bundle dataset, where the strongest neural baseline, Atten-Mixer, achieves an HR@1 of 0.0630 compared to \method's 0.3263. On ML-1M, CGM-Rec achieves the highest mean across all three reported metrics. On Games, it achieves the highest mean HR@1 and NDCG@5, while remaining competitive with ReLoop2 on HR@5; the differences on Games and ML-1M are not statistically significant. Furthermore, \method outperforms advanced LLM baselines, improving HR@1 by 29.58\% relative to LLMGreenRec on Bundle ($p<0.01$). This demonstrates that language-driven reasoning is insufficient for robust context adaptation, whereas structured, quality-gated graph mutations offer a critical architectural advantage. These findings validate our central hypothesis: adaptive recommenders achieve superior resilience by managing graph knowledge as an active, continually maintained memory layer rather than a fixed retrieval source.
 % \vspace{-1em}
\subsection{LLM-based Comparison in Metadata-rich}

Table~\ref{tab:llm_comparison} shows that \method outperforms the strongest LLM graph-augmented baseline, K-RagRec, by $+0.0971$ in HR@1 on the non-session, metadata-rich ML-100K setting, with all gains significant ($p<0.05$). This margin stems from K-RagRec's architectural limitation: it injects static, multi-hop subgraphs directly into the prompt as dense context. Large metadata overloads the context window with unstructured textual noise, obscuring subtle entity-attribute insights. In contrast, \method maps metadata taxonomies onto its structural layers, using a quality-gated write policy to filter and compact relation paths into a dual-timescale memory. This dynamic maintenance distills metadata into a noise-free retrieval state, grounding the LLM reranker in precise semantic neighborhoods even without sequential triggers. 

Candidate-pool scaling against K-RagRec at
$C\in\{20,50,80\}$ further shows that CGM-Rec maintains higher mean HR@5 on both Games and ML-100K. (Appendix~\ref{sec:appendix_candidate_scaling}).
% These findings validate that handling complex metadata requires treating graph knowledge as an active, continually maintained memory layer rather than an uncalibrated retrieval source.

\begin{table}[t]
\centering
\small
\setlength{\tabcolsep}{4pt}
\renewcommand{\arraystretch}{1.1}
\resizebox{\columnwidth}{!}{%
\begin{tabular}{l cccc r}
\toprule
\textbf{Metric} & \textbf{NIR} & \textbf{LLMGreenRec} & \textbf{K-RagRec} & \textbf{\method} & \textbf{$p$-value} \\
\midrule
HR@1    & 0.0037$\pm$.0022 & 0.0152$\pm$.0022 & \underline{0.2023}$\pm$.0139 & \textbf{0.2994}$\pm$.0032 & 1.28e-04 \\
HR@5    & 0.0315$\pm$.0084 & 0.1318$\pm$.0083 & \underline{0.4746}$\pm$.0258 & \textbf{0.5941}$\pm$.0087 & 5.12e-04 \\
HR@10   & 0.1193$\pm$.0122 & 0.2718$\pm$.0142 & \underline{0.6801}$\pm$.0381 & \textbf{0.7567}$\pm$.0105 & 1.12e-02 \\
NDCG@1  & 0.0037$\pm$.0022 & 0.0152$\pm$.0022 & \underline{0.2023}$\pm$.0139 & \textbf{0.2994}$\pm$.0032 & 1.28e-04 \\
NDCG@5  & 0.0167$\pm$.0053 & 0.0713$\pm$.0052 & \underline{0.3318}$\pm$.0268 & \textbf{0.4508}$\pm$.0096 & 6.01e-04 \\
NDCG@10 & 0.0546$\pm$.0055 & 0.1160$\pm$.0061 & \underline{0.3877}$\pm$.0416 & \textbf{0.5015}$\pm$.0119 & 3.86e-03 \\
\bottomrule
\end{tabular}}
\caption{Recommendation performance comparison against LLM-based baselines on the non-session, metadata-rich ML-100K benchmark. Best results are in bold and runner-up results are underlined.}
\vspace{-1.8em}
\label{tab:llm_comparison}
\end{table}
 % \vspace{-1em}
\subsection{Dual-Memory Ablation Study}
To assess dual-memory complementarity, CGM-Rec is compared against SGM-only, ELM-only, and static variants. The full system achieves the best performance, confirming that episodic memory handles rapid intent shifts while semantic memory provides stable long-term relational evidence. Detailed ablation results are provided in Appendix~\ref{sec:appendix_ablation}. Quality-gate sensitivity is further examined in Appendix~\ref{sec:appendix_gate_sensitivity}, where moderate threshold perturbations produce only small HR@5 changes despite substantial variation in accepted-write rates.
 % \vspace{-1em}
\subsection{Continual Adaptation}

This analysis evaluates continual adaptation on the chronological Bundle test stream, after initialization with a 20\% warm-up subset of the training stream. After each session is ranked, the target item is revealed and its feedback is incorporated into the adaptive state, with memory updates applied every five sessions while model parameters remain frozen. At each evaluation snapshot, the current adaptive state is frozen and evaluated on the following test samples. The plotted curve reports the 20-checkpoint rolling mean of HR@5 over five matched seeds, with shaded regions indicating $\pm1$ standard deviation. Thus, each point reflects how well the system performs on future samples given the accumulated memory.

\begin{figure}[t]
  \centering
  \includegraphics[width=0.92\columnwidth]{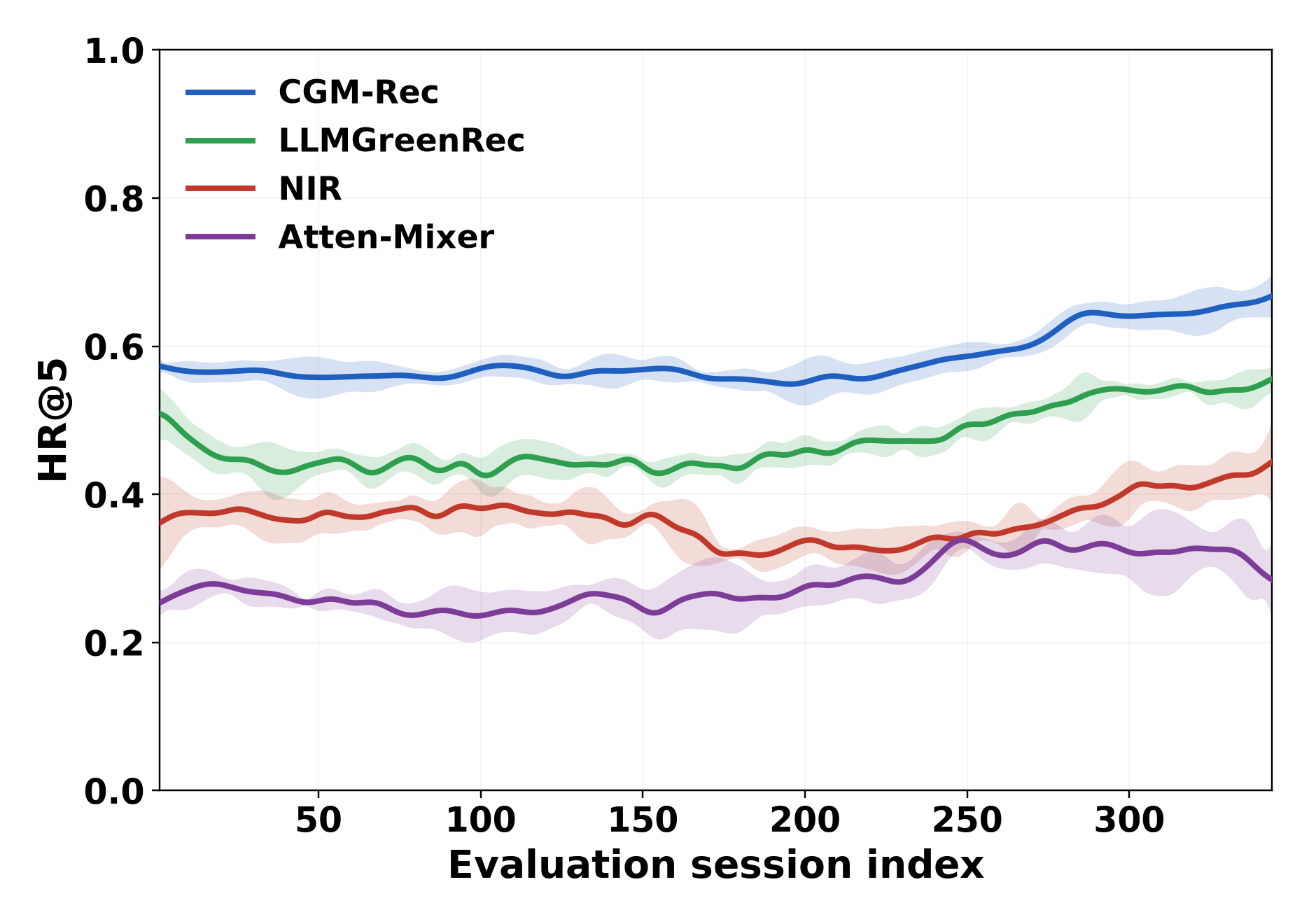}
\vspace{-0.8em}
\caption{Continual adaptation on Bundle using 20-checkpoint rolling-mean HR@5 over five seeds, with memory updated every five sessions and shaded regions indicating ±1 SD.}
\vspace{-1.2em}
\label{fig:model_comparison}
\end{figure}

Figure~\ref{fig:model_comparison} shows that CGM-Rec outperforms all baselines throughout the stream, with its advantage exceeding the $\pm1$ SD bands in later snapshots. Its curve also increases in later snapshots, indicating the system does not merely correct isolated failures but gradually accumulates feedback into memory. LLMGreenRec improves over time but remains below CGM-Rec, while NIR and Atten-Mixer show lower and less stable trajectories. These results highlight the continual nature of CGM-Rec: feedback from earlier sessions is converted into reusable graph memory and episodic lessons, enabling stronger adaptation.

\subsection{Adaptation under Intent-Regime Shifts}
\label{sec:intent_shift}

To connect measurable intent shifts with adaptation performance, a category-blocked regime-shift analysis is conducted on the Bundle test stream. Since Bundle does not preserve persistent user identities across sessions, the observed changes are treated as stream-level intent-regime shifts rather than latent within-user preference drift. Following prior work on temporal preference dynamics and concept drift \citep{koren2009collaborative,gama2014survey}, observable session intent is represented by the normalized distribution $\mathbf{p}_t$ of taxonomy labels among the context items. The shift magnitude at boundary $b$ is measured using Jensen--Shannon divergence (JSD) \citep{lin1991divergence} between the mean intent distributions of the 20 sessions before and after the boundary:
\begin{equation}
D_b =
\operatorname{JSD}
\left(
\overline{\mathbf{p}}_{b-20:b-1},
\overline{\mathbf{p}}_{b:b+19}
\right).
\end{equation}
Using base-2 logarithms, $D_b \in [0,1]$. The Bundle test stream contains three consecutive regimes: Food (78 sessions), Electronics (77), and Clothing (83), with JSD values of $1.000$ for Food$\rightarrow$Electronics and $0.979$ for Electronics$\rightarrow$Clothing. The drift score is used only for retrospective analysis and is never provided to the recommender or memory writer.

Adaptation is evaluated over five matched seeds using fixed 20-session windows before, immediately after, and in a later window after each shift. Full CGM-Rec, Static SGM+ELM, and K-RagRec use identical contexts, candidate pools, and stream order, while memory is not reset at regime boundaries. The HR@5 recovery ratio is computed from the aggregated means as follows:
\begin{equation}
\resizebox{0.94\columnwidth}{!}{$
\displaystyle
\mathrm{Recovery} =
\frac{\mathrm{HR@5}_{\mathrm{Late}}-\mathrm{HR@5}_{\mathrm{Immediate}}}
{\mathrm{HR@5}_{\mathrm{Before}}-\mathrm{HR@5}_{\mathrm{Immediate}}}
\times 100\%.
$}
\label{eq:recovery}
\end{equation}

\begin{table}[t]
\centering
\setlength{\tabcolsep}{2pt}
\resizebox{\columnwidth}{!}{%
\begin{tabular}{llcccc}
\toprule
Transition & Method & Before & Immediate & Later & Recovery \\
\midrule
Food$\rightarrow$Elec.
& \textbf{CGM-Rec} & \textbf{0.600 $\pm$ 0.0354} & \textbf{0.520 $\pm$ 0.0274} & \textbf{0.550 $\pm$ 0.0354} & \textbf{37.5\%} \\
& Static SGM+ELM & 0.570 $\pm$ 0.0274 & 0.480 $\pm$ 0.0274 & 0.490 $\pm$ 0.0224 & 11.1\% \\
& K-RagRec & 0.550 $\pm$ 0.0354 & 0.480 $\pm$ 0.0274 & 0.500 $\pm$ 0.0354 & 28.6\% \\
\midrule
Elec.$\rightarrow$Cloth.
& \textbf{CGM-Rec} & \textbf{0.550 $\pm$ 0.0354} & \textbf{0.480 $\pm$ 0.0274} & \textbf{0.510 $\pm$ 0.0418} & \textbf{42.9\%} \\
& Static SGM+ELM & 0.490 $\pm$ 0.0224 & 0.430 $\pm$ 0.0274 & 0.440 $\pm$ 0.0224 & 16.7\% \\
& K-RagRec & 0.500 $\pm$ 0.0354 & 0.440 $\pm$ 0.0224 & 0.460 $\pm$ 0.0418 & 33.3\% \\
\bottomrule
\end{tabular}}
\caption{HR@5 around intent-regime shifts on Bundle, reported as 20-session window means over five matched seeds rather than full-stream averages.}
\label{tab:intent_shift}
\end{table}

As shown in Table~\ref{tab:intent_shift}, all methods experience an immediate performance decline after each transition. CGM-Rec subsequently recovers a larger proportion of the lost HR@5 than Static SGM+ELM and K-RagRec. The recovery remains partial: CGM-Rec's later HR@5 is still $0.050$ and $0.040$ below the pre-shift level after the two transitions, respectively. These results strengthen the empirical connection between feedback-driven writable memory and post-shift adaptation under measurable intent shifts, rather than indicating immunity to drift or within-user preference recovery.

\vspace{-0.6em}
\section{Conclusion}
\vspace{-0.6em}
This paper presented \method, a continual graph-memory framework for adaptive recommendation under intent drift. The central idea is to move KG-enhanced recommendation beyond read-only graph retrieval by treating the graph state as a writable memory. \method separates fast feedback capture from stable semantic consolidation via Episodic Lesson Memory and Semantic Graph Memory, and controls graph changes with a quality-gated writer that converts recommendation outcomes into typed, provenance-aware edits. Under a frozen-parameter, one-pass sampled-candidate reranking protocol, \method achieves the strongest mean results on Bundle and remains competitive with strong neural and continual baselines on Games and ML-1M, while consistently outperforming K-RagRec in the metadata-rich ML-100K setting. Ablation and continual-stream analyses further suggest that episodic lessons and semantic graph memory provide complementary benefits. 

% These findings support the view that adaptive recommendation can benefit from maintaining graph evidence as long-term memory, rather than only retrieving from a static KG. 

% Future work will extend CGM-Rec to full-catalog ranking, richer user-level personalization, and more detailed cost, latency, and safety analyses.
 \vspace{-0.6em}
\section{Limitations}
 \vspace{-0.3em}

\method introduces additional complexity compared with static retrieval, requiring extra storage and computation for memory maintenance. Although the quality-gate sensitivity analysis shows robustness to moderate perturbations, broader gate configurations and typed update magnitudes remain unexplored. Evaluation focuses on sampled-candidate reranking, with the $C=50$ and
$C=80$ settings serving as larger-pool robustness tests rather than full-catalog evaluations. The controlled Bundle analysis captures stream-level intent-regime shifts rather than natural within-user preference drift. Finally, episodic-to-semantic promotion may propagate biased or noisy evidence, motivating stronger provenance safeguards.

% \section*{Ethics Statement}

% Continual recommendation systems can amplify harmful correlations if popularity-driven or biased patterns are repeatedly reinforced. Our design partially mitigates this risk by storing provenance, allowing suppression, and requiring repeated support before promotion into semantic memory. When user-level histories are available, memory construction must also respect privacy constraints, dataset licenses, and data minimization principles. The proposed framework is intended for offline research and should be audited for fairness, privacy, and robustness before deployment.

% \section*{Acknowledgments}

% Omitted for anonymous review.
% \bibliographystyle{acl_natbib}
\bibliography{latex/custom}

\appendix

\section{Relation-Aware Graph Encoder}
\label{app:graph_encoder}

\paragraph{Node initialization.}
Each node \(v\in V_0\) is initialized from a textual representation \(x_v\). For an item node, \(x_v\) concatenates available item metadata such as title, category, keywords, and description; for an attribute node, \(x_v\) is constructed from the corresponding attribute text. A frozen text encoder maps \(x_v\) into an initial normalized representation:
\begin{equation}
h_v^{(0)}
=
\frac{\mathrm{Enc}(x_v)}
{\|\mathrm{Enc}(x_v)\|_2}.
\end{equation}
The text encoder is not updated during graph training or test-time inference.

\paragraph{Relation-aware propagation.}
To incorporate typed graph structure, we use a relation-aware graph attention encoder over the seed graph \(G_0\). For an edge \((u,r,v)\), the attention weight from node \(u\) to neighbor \(v\) under relation \(r\) at layer \(l\) is:
\begin{equation}
\resizebox{0.94\columnwidth}{!}{$
a_{u,r,v}^{(l)}
=
\mathrm{softmax}_{(u,r',v')\in\mathcal{N}(u)}
\left(
\frac{(h_v^{(l)})^\top W_r^{(l)} h_u^{(l)}}{\tau}
\right),
$}
\end{equation}

% \begin{equation}
% a_{u,r,v}^{(l)}
% =
% \mathrm{softmax}_{(u,r',v')\in\mathcal{N}(u)}
% \left(
% \frac{(h_v^{(l)})^\top W_r^{(l)} h_u^{(l)}}{\tau}
% \right),
% \end{equation}
where \(W_r^{(l)}\) is a relation-specific transformation, \(\tau\) is a temperature parameter, and \(\mathcal{N}(u)\) denotes the typed neighborhood of \(u\). Node representations are updated by aggregating relation-weighted messages:
\begin{equation}
h_u^{(l+1)}
=
\sigma\left(
\sum_{(u,r,v)\in\mathcal{N}(u)}
a_{u,r,v}^{(l)} W_r^{(l)}h_v^{(l)}
\right).
\end{equation}
The final node representation concatenates multi-hop representations:
\begin{equation}
h_v^{*}=h_v^{(0)}\Vert h_v^{(1)}\Vert\cdots\Vert h_v^{(L)}.
\end{equation}

\paragraph{Offline training objective.}
The seed graph provides the structural substrate, but the graph encoder is trained offline using the training stream \(D_{\mathrm{train}}\). For each training instance \((x_t,C_t,y_t)\), the system extracts candidate-conditioned graph evidence from \(G_0\) around the context items and candidate items. For a candidate \(i\in C_t\), the encoder produces graph-based evidence features from the retrieved paths or edges, denoted by \(\phi_g(i\mid x_t,G_0)\). A lightweight scoring module then computes:
\begin{equation}
s_{\theta}(i\mid x_t,G_0)
=
\mathrm{Score}_{\theta}
\left(
h_i^{*}, h_{x_t}, \phi_g(i\mid x_t,G_0)
\right),
\end{equation}
where \(h_i^{*}\) is the graph representation of candidate item \(i\), and \(h_{x_t}\) is a context representation obtained from the items or textual attributes in \(x_t\).

The encoder and scoring module are optimized with a candidate-ranking loss over \(D_{\mathrm{train}}\):

\begin{equation}
\resizebox{\columnwidth}{!}{$
\mathcal{L}_{\mathrm{rank}}
=
-
\sum_{(x_t,C_t,y_t)\in D_{\mathrm{train}}}
\log
\frac{
\exp(s_{\theta}(y_t\mid x_t,G_0))
}{
\sum_{i\in C_t}
\exp(s_{\theta}(i\mid x_t,G_0))
}.
$}
\end{equation}

% \begin{equation}
% \small
% \mathcal{L}_{\mathrm{rank}}
% =
% -
% \sum_{(x_t,C_t,y_t)\in D_{\mathrm{train}}}
% \log
% \frac{
% \exp(s_{\theta}(y_t\mid x_t,G_0))
% }{
% \sum_{i\in C_t}
% \exp(s_{\theta}(i\mid x_t,G_0))
% }.
% \end{equation}
This objective encourages the encoder to assign higher graph-evidence scores to the ground-truth target item than to other candidates in the same candidate set.

\paragraph{Test-time retrieval.}
Before one-pass test inference, the graph encoder and scoring module are frozen. At test step \(t\), the current SGM state \(\mathcal{M}_{s,t}\) may differ from the initial seed graph because accepted memory edits can update edge weights, confidence scores, support counts, or add and prune semantic edges. Given the current context \(x_t\) and candidate set \(C_t\), the retriever extracts local candidate-conditioned subgraphs from \(\mathcal{M}_{s,t}\), including paths connecting context items, candidate items, and metadata nodes. The frozen encoder is then used only to score and rank these edges or paths; no encoder parameters are updated on the test stream. The top-ranked graph evidence is returned as \(R_t^s\) and used by the Recommender Agent.

\section{Quality Gate Details}
\label{app:quality_gate}

The quality gate \(U_{\phi}\) is a deterministic write filter, not a learned model. It uses the write signals in Eq.~\ref{eq:write_signals} to decide whether an Analyzer-generated proposal can update SGM. The gate can be expressed as a thresholded score:
\begin{equation}
\small
\begin{split}
\mathrm{WriteScore}_t(p)
=&\;
\theta_1 O_t(p)
+\theta_2 Q_t(p) \\
&+\theta_3 \mathrm{Support}_t(p)
+\theta_4 \mathrm{Recency}_t(p) \\
&-\theta_5 \mathrm{Conflict}_t(p)
-\theta_6 \mathrm{Cost}_t(p).
\end{split}
\end{equation}
Here, positive terms reward useful outcomes, proposal quality, repeated support, and recency, while negative terms penalize conflict with SGM and memory cost. A proposal is accepted if
\begin{equation}
\mathrm{Accept}(p)=
\mathbb{I}
[
\mathrm{WriteScore}_t(p)\geq \eta_a
].
\end{equation}
All coefficients and thresholds are included in \(\phi\), fixed before inference, and not tuned on the test stream. Action-specific rules constrain the decision: reinforcement requires positive evidence, suppression requires repeated misleading evidence, promotion requires repeated support and low conflict, and pruning targets stale or low-utility relations.

\section{SGM Update Rules}
\label{app:sgm_update}

The \(\mathrm{Apply}(\cdot)\) operator updates SGM by executing the accepted typed actions in \(\mathcal{A}_t\). Each edge \(e=(u,r,v)\) maintains a support count \(n_{e,t}\), confidence \(c_{e,t}\), and weight \(w_{e,t}\). These attributes are updated conservatively so that a single Analyzer proposal cannot immediately create or remove stable semantic knowledge.

For a \texttt{reinforce\_edge} action, the edge receives additional support and its reliability is increased:
\begin{align}
n_{e,t+1} &= n_{e,t}+1,\\
c_{e,t+1} &= \min(1,c_{e,t}+\Delta c),\\
w_{e,t+1} &= w_{e,t}+\Delta w.
\end{align}
Here, \(\Delta c>0\) and \(\Delta w>0\) are fixed update steps for confidence and edge weight. Confidence is clipped to the range \([0,1]\), while the weight controls the strength of the relation during later retrieval.

For a \texttt{suppress\_edge} action, the edge is treated as potentially misleading and its reliability is reduced:
\begin{align}
c_{e,t+1} &= \max(0,c_{e,t}-\Delta c),\\
w_{e,t+1} &= \max(0,w_{e,t}-\Delta w).
\end{align}
This operation does not necessarily delete the edge immediately. Instead, it weakens relations that repeatedly conflict with observed outcomes or retrieve misleading evidence.

For \texttt{add\_tentative\_edge}, a new relation is inserted with low initial confidence and limited weight. Such an edge must accumulate repeated support before it can be promoted or treated as stable semantic knowledge. This prevents a single noisy observation from becoming a permanent graph relation.

Graph compaction is handled through deterministic pruning or deactivation. An edge is removed or deactivated when all of its utility indicators fall below predefined minimum thresholds:
\[
w_{e,t}<\omega_{\min}
\;\land\;
c_{e,t}<c_{\min}
\;\land\;
n_{e,t}<n_{\min}.
\]
This rule removes relations that are simultaneously weak, low-confidence, and insufficiently supported. It keeps SGM compact and limits the accumulation of noisy or stale edges during long one-pass evaluation streams. All update steps and pruning thresholds are fixed before test-time inference.

\begin{table*}[t]
\centering
\small
\setlength{\tabcolsep}{3.5pt}
\renewcommand{\arraystretch}{1.08}
\begin{tabular}{lrrrrcc}
\toprule
\textbf{Dataset}
& \textbf{Source interactions}
& \textbf{Eligible items}
& \textbf{Train episodes}
& \textbf{Test episodes}
& \textbf{Train context}
& \textbf{Test context} \\
&
&
&
&
&
\textbf{(min/mean/max)}
&
\textbf{(min/mean/max)} \\
\midrule

Bundle
& 18,886$^{*}$
& 14,240
& 146
& 238
& 2 / 5.53 / 9
& 3 / 6.05 / 9 \\

Games
& 497,577
& 17,389
& 141
& 1,000
& 1 / 3.25 / 10
& 1 / 3.61 / 9 \\

ML-1M
& 1,000,209
& 3,416
& 150
& 1,000
& 19 / 19.00 / 19
& 1 / 18.01 / 19 \\

ML-100K
& 100,000
& 1,500
& 200
& 743
& 19 / 42.74 / 50
& 19 / 41.73 / 50 \\

\bottomrule
\end{tabular}
\caption{Dataset and processed recommendation-stream statistics. Source interactions denote benchmark events before episode construction, while eligible items denote the post-filter ranking vocabulary. Train and test counts denote processed recommendation episodes rather than source interactions. Context length is reported as minimum/mean/maximum. $^{*}$The Bundle source release reports 18,886 interactions; the artifact-local \texttt{user\_item} snapshot contains 16,966 rows.}
\label{tab:data_stats}
\end{table*}

% \caption{Dataset statistics. The table summarizes the number of instances and the distribution of context lengths (minimum, average, and maximum) for both training and testing phases across the four evaluated datasets.}

\section{Implementation Hyperparameters}
\label{app:hyperparameters}

Table~\ref{tab:implementation_config} reports the default implementation configuration used in our experiments. Unless otherwise stated, the same values are shared across datasets. These settings are fixed before one-pass test-time inference and are not tuned on the test stream. During evaluation, the graph scorer, and prompt templates remain frozen; adaptation occurs only through SGM and ELM updates.

\begin{table*}[t]
\centering
\small
\setlength{\tabcolsep}{5pt}
\renewcommand{\arraystretch}{1.1}
\begin{tabularx}{\textwidth}{lX}
\toprule
\textbf{Component} & \textbf{Default configuration} \\
\midrule
Data / stream setup 
& Candidate size = 20 for all compared methods; random seed = 42; seed initialization = 20\% of the training stream. \\

SGM seed graph 
& Keyword top-$k$ = 5; co-occurrence window = 5; item descriptions enabled. \\

Graph scorer 
& Hidden dimension = 32; max edges per seed node = 64; optimizer = Adam. \\

ELM retrieval 
& Retrieved lessons top-$k$ = 5; BM25 $k_1 = 1.2$, $b = 0.75$. \\

Analyzer feedback 
& Success threshold = Top-5; max success edits = 4; max failure edits = 6; reserved tentative edits = 2. \\

Typed edit deltas 
& Reinforce $(\Delta w,\Delta c,\Delta q)=(0.20,0.05,0.03)$; suppress $=(-0.15,-0.04,-0.03)$; tentative $=(0.10,0.02,0.02)$. \\

Quality gate 
& Reinforce threshold = 0.35; suppress threshold = 0.45; tentative threshold = 0.30; contribution scale = 0.25; support scale = 5.0; max conflict penalty = 0.75. \\

Semantic promotion 
& Min support = 3; min confidence = 0.50; min quality = 0.45; min unique contexts = 2; check interval = 20 steps. \\

LLM output control 
& JSON response format; max generation attempts = 3. \\
\bottomrule
\end{tabularx}
\caption{Default implementation configuration for CGM-Rec. Unless otherwise stated, the same settings are used across datasets, fixed before one-pass test-time inference, and not tuned on the test stream.}
\label{tab:implementation_config}
\end{table*}

\section{Runtime and Token Overhead}
\label{sec:appendix_efficiency}
Table~\ref{tab:inference_overhead} shows that the two LLM calls dominate runtime, while memory retrieval and update add approximately 0.2\,s per instance. Token usage varies across instances with context and retrieved memory. Overall, the measurements highlight the practical trade-off of the dual-agent design: CGM-Rec avoids test-time parameter updates but requires two LLM calls per recommendation.

\begin{table}[t]
\centering
\setlength{\tabcolsep}{3.5pt}
\resizebox{\columnwidth}{!}{%
\begin{tabular}{lccc}
\toprule
\textbf{Component}
& \textbf{Avg. Latency (s)}
& \textbf{Avg. Input}
& \textbf{Avg. Output} \\
\midrule
Recommender
& 4.6
& 9.66K
& 0.78K \\

Analyzer
& 4.8
& 10.27K
& 0.39K \\

Memory retrieval/update
& 0.2
& -- 
& -- \\

\midrule
\textbf{Total}
& \textbf{9.6}
& \textbf{19.93K}
& \textbf{1.17K} \\
\bottomrule
\end{tabular}}
\caption{Average per-instance inference overhead of CGM-Rec. Input and output columns report token counts; memory operations do not invoke the LLM.}
\label{tab:inference_overhead}
\end{table}

\section{Full-Catalog for Neural Baselines}
\label{sec:appendix_full_catalog}
To verify the fairness of the shared sampled-candidate protocol, NARM, STAMP, GCE-GNN, and Atten-Mixer are additionally evaluated on ML-1M under both sampled-20 and full-catalog ranking. The processed evaluation catalog contains 3,416 eligible items. Full-catalog evaluation ranks all eligible items, whereas sampled-20 evaluation
restricts the same model score vector to the shared 20 candidates. No retraining or parameter tuning is performed between protocols; test contexts, item mappings, scoring functions, eligibility masks, and tie-breaking rules remain unchanged.

\begin{table}[t]
\centering
\small
\setlength{\tabcolsep}{3.5pt}
\begin{tabular}{lcccc}
\toprule
Model & HR@1 & HR@5 & NDCG@5 & Med. Rank $\downarrow$ \\
\midrule
NARM        & 0.0000 & 0.0010 & 0.0004 & 1483.0 \\
STAMP       & 0.0040 & 0.0080 & 0.0060 & 747.0 \\
GCE-GNN     & 0.0010 & 0.0070 & 0.0043 & 1029.0 \\
Atten-Mixer & 0.0020 & 0.0070 & 0.0044 & 828.5 \\
\bottomrule
\end{tabular}
\caption{Full-catalog sanity check for neural baselines on ML-1M with 3,416 eligible items.}
\label{tab:full_catalog_neural}
\end{table}

As expected, absolute ranking scores decrease under the 3,416-item catalog. Restricting each full-catalog score vector to its 20 candidates reproduces the exact sampled-20 target ranks. This consistency check confirms that the neural baselines produce candidate-independent catalog scores and that the shared 20-item protocol compares all methods on an identical reranking space. Full-catalog numbers serve only as a neural baseline sanity check, not for direct comparison with \method.

\section{Detailed Ablation Study (RQ3)}
\label{sec:appendix_ablation}
Table~\ref{tab:ablation_components} separates two complementary effects in \method: dual-memory composition and feedback-driven semantic-graph updating. The dual-memory contribution is reflected by the difference between Static SGM+ELM and the stronger single-memory variant, whereas the write contribution is reflected by the difference between Full CGM-Rec and Static SGM+ELM. Across datasets, the static dual-memory configuration yields higher mean performance than either memory alone, while enabling feedback-driven graph updates provides a further mean gain whose magnitude varies by dataset. The write contribution is modest on Bundle and Games, more noticeable on ML-1M, and most pronounced on ML-100K, where HR@5 increases from $0.4460$ to $0.5941$. Overall, these
results indicate that dual-memory retrieval contributes beyond either memory alone, while the complete feedback-driven write pathway provides an additional, dataset-dependent source of adaptation.
\begin{table*}[t]
\centering
\small
\setlength{\tabcolsep}{4pt}
\renewcommand{\arraystretch}{1.1}
\resizebox{\textwidth}{!}{%
\begin{tabular}{ll ccc ccc}
\toprule
\textbf{Dataset} & \textbf{Method}
& \textbf{HR@1} & \textbf{HR@5} & \textbf{HR@10}
& \textbf{NDCG@1} & \textbf{NDCG@5} & \textbf{NDCG@10} \\
\midrule

\multirow{4}{*}{Bundle}
& SGM only
& $0.2748 \pm 0.0125$
& $0.5486 \pm 0.0148$
& $0.6842 \pm 0.0163$
& $0.2748 \pm 0.0125$
& $0.4201 \pm 0.0127$
& $0.4702 \pm 0.0139$ \\

& ELM only
& $0.2827 \pm 0.0141$
& $0.5538 \pm 0.0155$
& $0.7105 \pm 0.0149$
& $0.2827 \pm 0.0141$
& $0.4246 \pm 0.0133$
& $0.4753 \pm 0.0142$ \\

& Static SGM+ELM
& $0.3101 \pm 0.0154$
& $0.5634 \pm 0.0126$
& $0.7168 \pm 0.0138$
& $0.3101 \pm 0.0154$
& $0.4467 \pm 0.0131$
& $0.4859 \pm 0.0137$ \\

& \textbf{Full CGM-Rec}
& $\mathbf{0.3263 \pm 0.0175}$
& $\mathbf{0.5716 \pm 0.0098}$
& $\mathbf{0.7290 \pm 0.0125}$
& $\mathbf{0.3263 \pm 0.0175}$
& $\mathbf{0.4495 \pm 0.0146}$
& $\mathbf{0.5060 \pm 0.0140}$ \\
\midrule

\multirow{4}{*}{Games}
& SGM only
& $0.2990 \pm 0.0128$
& $0.5864 \pm 0.0160$
& $0.6985 \pm 0.0180$
& $0.2990 \pm 0.0128$
& $0.4435 \pm 0.0152$
& $0.4760 \pm 0.0160$ \\

& ELM only
& $0.3021 \pm 0.0132$
& $0.5848 \pm 0.0165$
& $0.7106 \pm 0.0171$
& $0.3021 \pm 0.0132$
& $0.4426 \pm 0.0156$
& $0.4790 \pm 0.0162$ \\

& Static SGM+ELM
& $0.3050 \pm 0.0129$
& $0.5877 \pm 0.0168$
& $0.7138 \pm 0.0165$
& $0.3050 \pm 0.0129$
& $0.4458 \pm 0.0160$
& $0.4822 \pm 0.0158$ \\

& \textbf{Full CGM-Rec}
& $\mathbf{0.3092 \pm 0.0131}$
& $\mathbf{0.5898 \pm 0.0176}$
& $\mathbf{0.7240 \pm 0.0160}$
& $\mathbf{0.3092 \pm 0.0131}$
& $\mathbf{0.4475 \pm 0.0168}$
& $\mathbf{0.4868 \pm 0.0156}$ \\
\midrule

\multirow{4}{*}{ML-1M}
& SGM only
& $0.1327 \pm 0.0078$
& $0.5086 \pm 0.0138$
& $0.6812 \pm 0.0125$
& $0.1327 \pm 0.0078$
& $0.3250 \pm 0.0098$
& $0.3810 \pm 0.0105$ \\

& ELM only
& $0.1410 \pm 0.0081$
& $0.5500 \pm 0.0132$
& $0.7515 \pm 0.0121$
& $0.1410 \pm 0.0081$
& $0.3545 \pm 0.0107$
& $0.4180 \pm 0.0112$ \\

& Static SGM+ELM
& $0.1675 \pm 0.0093$
& $0.5572 \pm 0.0130$
& $0.7582 \pm 0.0118$
& $0.1675 \pm 0.0093$
& $0.3667 \pm 0.0114$
& $0.4300 \pm 0.0117$ \\

& \textbf{Full CGM-Rec}
& $\mathbf{0.2049 \pm 0.0117}$
& $\mathbf{0.5786 \pm 0.0174}$
& $\mathbf{0.7715 \pm 0.0142}$
& $\mathbf{0.2049 \pm 0.0117}$
& $\mathbf{0.3878 \pm 0.0156}$
& $\mathbf{0.4395 \pm 0.0138}$ \\
\midrule

\multirow{4}{*}{ML-100K}
& SGM only
& $0.1185 \pm 0.0105$
& $0.3220 \pm 0.0185$
& $0.5790 \pm 0.0200$
& $0.1185 \pm 0.0105$
& $0.2210 \pm 0.0148$
& $0.3050 \pm 0.0160$ \\

& ELM only
& $0.1372 \pm 0.0118$
& $0.4050 \pm 0.0190$
& $0.6690 \pm 0.0182$
& $0.1372 \pm 0.0118$
& $0.2695 \pm 0.0165$
& $0.3520 \pm 0.0171$ \\

& Static SGM+ELM
& $0.1538 \pm 0.0102$
& $0.4460 \pm 0.0160$
& $0.6765 \pm 0.0150$
& $0.1538 \pm 0.0102$
& $0.2978 \pm 0.0138$
& $0.3655 \pm 0.0145$ \\

& \textbf{Full CGM-Rec}
& $\mathbf{0.2994 \pm 0.0032}$
& $\mathbf{0.5941 \pm 0.0087}$
& $\mathbf{0.7567 \pm 0.0105}$
& $\mathbf{0.2994 \pm 0.0032}$
& $\mathbf{0.4508 \pm 0.0096}$
& $\mathbf{0.5015 \pm 0.0119}$ \\

\bottomrule
\end{tabular}}
\caption{Ablation results reported as mean$\pm$SD over five matched seeds. Static SGM+ELM retains both memory components while disabling feedback-driven semantic-graph updates.}
\label{tab:ablation_components}
\end{table*}

\section{Quality-Gate Sensitivity Analysis}
\label{sec:appendix_gate_sensitivity}

To examine whether CGM-Rec depends on narrowly tuned quality-gate parameters, a one-factor-at-a-time sensitivity analysis is conducted on ML-100K. The action-specific acceptance threshold $\eta_{\mathrm{sup}}$ and the outcome-signal coefficient $\theta_{\mathrm{out}}$ are varied independently, while all other gate parameters, typed update rules, prompts, candidate sets, memory budgets, and initial memory states remain fixed. The default configuration is shown in bold.

\begin{table}[t]
\centering
\small
\setlength{\tabcolsep}{4pt}
\begin{tabular}{ccccc}
\toprule
$\eta_{\mathrm{sup}}$ & HR@1 & HR@5 & NDCG@5 & Accept. (\%) \\
\midrule
0.25 & 0.2968 & 0.5970 & 0.4564 & 91.8 \\
0.35 & 0.2992 & 0.6014 & 0.4600 & 89.8 \\
\textbf{0.45} & \textbf{0.3005} & \textbf{0.6032} &
\textbf{0.4618} & \textbf{87.3} \\
0.55 & 0.3011 & 0.6040 & 0.4625 & 81.7 \\
0.65 & 0.2961 & 0.5965 & 0.4553 & 70.6 \\
\bottomrule
\end{tabular}
\caption{Sensitivity to the acceptance threshold
$\eta_{\mathrm{sup}}$ on ML-100K. "Accept.\" denotes the percentage of
proposed typed updates accepted by the quality gate.}
\label{tab:sensitivity_eta}
\end{table}

\begin{table}[t]
\centering
\small
\setlength{\tabcolsep}{4pt}
\begin{tabular}{ccccc}
\toprule
$\theta_{\mathrm{out}}$ & HR@1 & HR@5 & NDCG@5 & Accept. (\%) \\
\midrule
0.10 & 0.2955 & 0.5958 & 0.4540 & 72.5 \\
0.15 & 0.2987 & 0.6009 & 0.4592 & 80.6 \\
\textbf{0.20} & \textbf{0.3005} & \textbf{0.6032} &
\textbf{0.4618} & \textbf{87.3} \\
0.25 & 0.3012 & 0.6038 & 0.4622 & 91.3 \\
0.30 & 0.2972 & 0.5988 & 0.4570 & 94.8 \\
\bottomrule
\end{tabular}
\caption{Sensitivity to the outcome-signal coefficient
$\theta_{\mathrm{out}}$ on ML-100K. The default configuration is
shown in bold.}
\label{tab:sensitivity_theta}
\end{table}

As shown in Tables~\ref{tab:sensitivity_eta} and~\ref{tab:sensitivity_theta}, varying either gate parameter substantially changes the accepted-update rate while recommendation performance remains relatively stable. Since accepted updates affect future states and retrieval, the analysis captures end-to-end sensitivity of the quality-gated update mechanism. The
default configuration is retained without post-hoc selection, and the results indicate robustness to moderate gate perturbations rather than arbitrary parameter settings.

\section{Candidate-Pool Scaling Analysis}
\label{sec:appendix_candidate_scaling}
To evaluate robustness beyond the original 20-item reranking protocol, CGM-Rec is compared with K-RagRec under larger candidate pools on Games and ML-100K. Both methods use identical test contexts, candidate files, instance order, and five matched candidate-set seeds $\{0,10,42,625,2023\}$. The candidate sets are nested as $C_{20}\subset C_{50}\subset C_{80}$, such that larger pools preserve the original candidates while introducing additional harder negatives.

\begin{table*}[t]
\centering
\small
\setlength{\tabcolsep}{5pt}
\renewcommand{\arraystretch}{1.08}

\begin{tabular}{cc l c c c c}
\toprule
$\mathbf{C}$ & \textbf{Metric}
& \textbf{K-RagRec}
& \textbf{CGM-Rec}
& \textbf{Rel. Gain}
& $\mathbf{p}$\textbf{-value} \\
\midrule

\multicolumn{6}{l}{\textbf{Games}} \\
\midrule

\multirow{3}{*}{20}
& HR@1   & $0.2850 \pm 0.0110$ & $\mathbf{0.3092 \pm 0.0131}$ & $+8.5\%$  & 0.0039 \\
& HR@5   & $0.5550 \pm 0.0139$ & $\mathbf{0.5898 \pm 0.0176}$ & $+6.3\%$  & 0.0028 \\
& NDCG@5 & $0.4140 \pm 0.0123$ & $\mathbf{0.4475 \pm 0.0168}$ & $+8.1\%$  & 0.0139 \\
\cmidrule(lr){1-6}

\multirow{3}{*}{50}
& HR@1   & $0.2000 \pm 0.0104$ & $\mathbf{0.2385 \pm 0.0104}$ & $+19.2\%$ & $<0.001$ \\
& HR@5   & $0.4200 \pm 0.0146$ & $\mathbf{0.4747 \pm 0.0106}$ & $+13.0\%$ & $<0.001$ \\
& NDCG@5 & $0.3200 \pm 0.0128$ & $\mathbf{0.3624 \pm 0.0107}$ & $+13.3\%$ & $<0.001$ \\
\cmidrule(lr){1-6}

\multirow{3}{*}{80}
& HR@1   & $0.1600 \pm 0.0104$ & $\mathbf{0.1989 \pm 0.0063}$ & $+24.3\%$ & 0.0045 \\
& HR@5   & $0.3600 \pm 0.0145$ & $\mathbf{0.4253 \pm 0.0075}$ & $+18.1\%$ & 0.0017 \\
& NDCG@5 & $0.2700 \pm 0.0125$ & $\mathbf{0.3177 \pm 0.0070}$ & $+17.7\%$ & 0.0034 \\

\midrule
\multicolumn{6}{l}{\textbf{ML-100K}} \\
\midrule

\multirow{3}{*}{20}
& HR@1   & $0.2023 \pm 0.0139$ & $\mathbf{0.2994 \pm 0.0032}$ & $+48.0\%$ & $<0.001$ \\
& HR@5   & $0.4746 \pm 0.0258$ & $\mathbf{0.5941 \pm 0.0087}$ & $+25.2\%$ & $<0.001$ \\
& NDCG@5 & $0.3318 \pm 0.0268$ & $\mathbf{0.4508 \pm 0.0096}$ & $+35.9\%$ & $<0.001$ \\
\cmidrule(lr){1-6}

\multirow{3}{*}{50}
& HR@1   & $0.1276 \pm 0.0197$ & $\mathbf{0.1491 \pm 0.0044}$ & $+16.9\%$ & 0.1067 \\
& HR@5   & $0.3327 \pm 0.0263$ & $\mathbf{0.3833 \pm 0.0060}$ & $+15.2\%$ & 0.0216 \\
& NDCG@5 & $0.2259 \pm 0.0237$ & $\mathbf{0.2683 \pm 0.0055}$ & $+18.8\%$ & 0.0277 \\
\cmidrule(lr){1-6}

\multirow{3}{*}{80}
& HR@1   & $0.0809 \pm 0.0219$ & $\mathbf{0.1098 \pm 0.0040}$ & $+35.7\%$ & 0.0610 \\
& HR@5   & $0.2407 \pm 0.0281$ & $\mathbf{0.2886 \pm 0.0066}$ & $+19.9\%$ & 0.0312 \\
& NDCG@5 & $0.1620 \pm 0.0260$ & $\mathbf{0.2026 \pm 0.0052}$ & $+25.1\%$ & 0.0388 \\

\bottomrule
\end{tabular}
\caption{Candidate-pool scaling on Games and ML-100K, reported as mean$\pm$SD over five matched candidate-set seeds. Larger candidate sets are nested and preserve the original candidates while introducing additional harder negatives. Relative gains are computed with respect to K-RagRec; $p$-values are obtained from paired two-sided $t$-tests across matched seeds.}
\label{tab:candidate_scaling}
\end{table*}

Table~\ref{tab:candidate_scaling} shows the expected decline in absolute ranking performance as the candidate pool becomes larger and more challenging. Nevertheless, CGM-Rec maintains higher mean performance than K-RagRec across all evaluated metrics and pool sizes. For HR@5, the relative gain on Games increases from $6.3\%$ at $C=20$ to $18.1\%$ at $C=80$, while ML-100K retains gains of $15.2$--$25.2\%$ across the tested pools. The HR@5 improvements are statistically significant at every candidate size on both datasets. These results indicate that the advantage of CGM-Rec persists beyond the original 20-item setting as the sampled reranking space grows. This analysis evaluates robustness to larger sampled candidate pools rather than full-catalog recommendation or training-scale scalability.

\section{Discussion}
\label{app:discussion}

The results suggest that \method gains primarily come from explicit memory maintenance rather than test-time parameter updates. The graph encoder, LLM agents, and prompt templates remain fixed during testing; only SGM and ELM are updated. Ablations further show that the two memories are complementary: ELM reuses recent outcome-derived lessons, while SGM preserves stable relational knowledge through conservative, quality-gated edits.

The evaluation should be interpreted within its intended scope. The examples are anonymous or weakly linked, so adaptation targets transferable graph knowledge across contexts rather than lifelong per-user modeling. The candidate sets are sampled, so the results measure constrained reranking rather than full-catalog recommendation. ML-100K is treated separately as a metadata-rich contextual reranking setting rather than as a natural session stream. These constraints motivate future work on full-catalog ranking, harder negative sampling, and broader domains.

\section*{Ethical Considerations}

We use established public recommendation benchmarks and do not collect new user data. The processed inputs contain no personally identifying information and are restricted to interaction/context signals and item metadata required for recommendation. Continual recommendation systems may amplify popularity bias, spurious correlations, or harmful feedback loops when biased outcomes are repeatedly written into memory. \method mitigates this risk through structured evidence, conservative quality-gated updates, suppression of misleading relations, and repeated support before semantic promotion. When user-level histories or sensitive attributes are available, memory construction must respect privacy constraints, dataset licenses, and data minimization principles. This work is intended for offline research evaluation; deployment requires further auditing for fairness, privacy, robustness, and memory-induced bias.

% \section{Implementation Notes}

% \paragraph{LLM usage.}
% If an LLM is used, we keep the ranking instruction fixed during \(\Dtest\). The LLM may consume the current context, candidate list, semantic graph evidence, and episodic lessons, but persistent updates must always be normalized into typed graph edits.

% \paragraph{Candidate generation.}
% All methods share the same dataset-provided candidate lists. If the candidate lists are sampled, the sampling procedure is kept fixed across methods and never uses future information.

% \paragraph{Memory budget.}
% The episodic memory budget can be implemented as a top-\(B\) reservoir scored by recency, utility, and retrieval frequency. Semantic memory grows more slowly because promotion is conservative by design.

\section{Recommender Agent Prompt Template}
\label{app:rec_temp}

The Recommender Agent uses a fixed instruction template across datasets. At inference time, dataset-specific fields are filled with the current interaction context, candidate items, retrieved SGM evidence, and retrieved ELM lessons. The same core instruction is used across domains, while available item attributes such as category, genre, taxonomy, keywords, or descriptions are inserted when applicable.

\begin{tcolorbox}[
    enhanced,
    breakable,
    colback=white,
    colframe=black,
    title={Prompt Template: Recommender Agent},
    coltitle=white,
    colbacktitle=black,
    fonttitle=\bfseries,
    fontupper=\small,
    sharp corners,
    boxrule=0.8pt,
    left=4pt,
    right=4pt,
    top=4pt,
    bottom=4pt,
    before skip=4pt,
    after skip=4pt
]
\textbf{Task Instruction.}
This is a recommendation reranking task. Analyze the user's current interaction context and the candidate set to infer the user's preferences and current intent. Use retrieved graph evidence and episodic lessons to support the reranking decision.

\medskip
\noindent\textbf{Inputs}
\begin{enumerate}[leftmargin=1.4em,itemsep=2pt,topsep=2pt]
    \item \textbf{User interaction context:}\\
    \texttt{\{interaction\_context\}}

    \item \textbf{Candidate set} \textit{(rerank exactly these items):}\\
    \texttt{\{candidate\_set\}}

    \item \textbf{Retrieved graph evidence:}\\
    \texttt{\{graph\_evidence\}}

    \item \textbf{Retrieved episodic lessons:}\\
    \texttt{\{episodic\_lessons\}}
\end{enumerate}

\medskip
\noindent\textbf{Reasoning Instructions}
\begin{enumerate}[leftmargin=1.4em,itemsep=2pt,topsep=2pt]
    \item Identify preference patterns, item combinations, or intent cues from the current interaction context.
    \item Compare candidate items with the inferred intent using available attributes, such as category, genre, taxonomy, keywords, descriptions, price, or other dataset-specific metadata.
    \item Use graph evidence to identify relevant semantic relations, item--attribute links, or item--item associations.
    \item Use episodic lessons to account for previous success patterns, failure cases, or corrective hints from similar contexts.
    \item Rerank all candidate items by their likelihood of matching the user's current intent.
\end{enumerate}

\medskip
\noindent\textbf{Critical Rules}
\begin{itemize}[leftmargin=1.4em,itemsep=2pt,topsep=2pt]
    \item Do not introduce items outside the Candidate Set.
    \item Return all candidate items ordered by relevance.
    \item The recommendations field must contain only candidate items and their corresponding indices.
    \item Return strictly in JSON format with no markdown.
\end{itemize}

\medskip
\noindent\textbf{Output Format}
\begin{quote}
\footnotesize\ttfamily
\{\\
\quad "reasoning": "short analysis of user intent and ranking evidence",\\
\quad "recommendations": [\\
\quad\quad \{"index": 1, "item": "item name 1"\},\\
\quad\quad \{"index": 2, "item": "item name 2"\}\\
\quad ]\\
\}
\end{quote}
\end{tcolorbox}

\section{Analyzer Agent Prompt Template}
\label{app:anal_temp}

The Analyzer Agent is invoked after the recommendation outcome is observed. It receives an outcome package containing the generated ranking, target item, outcome label, retrieved SGM evidence, and retrieved ELM lessons. The Analyzer produces reusable textual lessons and structured memory-edit proposals. These outputs are treated only as candidates and must pass the quality-gated write policy before any modification is applied to SGM.

\begin{tcolorbox}[
    enhanced,
    breakable,
    colback=white,
    colframe=black,
    title={Prompt Template: Analyzer Agent},
    coltitle=white,
    colbacktitle=black,
    fonttitle=\bfseries,
    fontupper=\small,
    sharp corners,
    boxrule=0.8pt,
    left=4pt,
    right=4pt,
    top=4pt,
    bottom=4pt,
    before skip=4pt,
    after skip=4pt
]
\textbf{Task Instruction.}
This is a post-feedback analysis task for a recommendation system. Analyze the recommendation outcome, identify useful or misleading evidence, extract reusable lessons, and propose possible memory-edit actions. All proposed updates are candidate actions and will be checked by a quality gate.

\medskip
\noindent\textbf{Inputs}
\begin{enumerate}[leftmargin=1.4em,itemsep=2pt,topsep=2pt]
    \item \textbf{User interaction context:}\\
    \texttt{\{interaction\_context\}}

    \item \textbf{Candidate set:}\\
    \texttt{\{candidate\_set\}}

    \item \textbf{Generated ranking:}\\
    \texttt{\{predicted\_ranking\}}

    \item \textbf{Target item and rank:}\\
    \texttt{\{target\_item\}}, \texttt{\{target\_rank\}}

    \item \textbf{Outcome label:}\\
    \texttt{\{outcome\_label\}}

    \item \textbf{Retrieved graph evidence:}\\
    \texttt{\{graph\_evidence\}}

    \item \textbf{Retrieved episodic lessons:}\\
    \texttt{\{episodic\_lessons\}}
\end{enumerate}

\medskip
\noindent\textbf{Analysis Instructions}
\begin{enumerate}[leftmargin=1.4em,itemsep=2pt,topsep=2pt]
    \item Check whether the generated ranking matches the target item and current user intent.
    \item Identify helpful evidence, such as useful item--attribute relations, item--item links, or intent cues.
    \item Identify misleading evidence, such as broad categories, weak relations, noisy attributes, or irrelevant lessons.
    \item Generate concise textual lessons for similar future contexts.
    \item Propose structured edit actions only when supported by the outcome package.
    \item Do not invent unsupported items, attributes, relations, or evidence.
\end{enumerate}

\medskip
\noindent\textbf{Allowed Candidate Outputs}
\begin{itemize}[leftmargin=1.4em,itemsep=2pt,topsep=2pt]
    \item \texttt{store\_lesson}: store a textual lesson in ELM.
    \item \texttt{add\_tentative\_edge}: propose a low-confidence candidate relation for SGM.
    \item \texttt{reinforce\_edge}: strengthen a useful existing relation.
    \item \texttt{suppress\_edge}: downweight a misleading relation.
    \item \texttt{promote\_to\_semantic}: promote a reliable recurring pattern into SGM.
    \item \texttt{prune\_stale\_edge}: remove or deactivate a stale low-utility relation.
\end{itemize}

\medskip
\noindent\textbf{Critical Rules}
\begin{itemize}[leftmargin=1.4em,itemsep=2pt,topsep=2pt]
    \item Output only candidate lessons and candidate edit proposals.
    \item Every edit proposal must include supporting evidence from the outcome package.
    \item If evidence is insufficient, return an empty list of edit proposals.
    \item Return strictly in JSON format with no markdown.
\end{itemize}

\medskip
\noindent\textbf{Output Format}
\begin{quote}
\scriptsize\ttfamily
\{\\
\quad "outcome\_summary": \{\\
\quad\quad "target\_rank": "\textless target rank\textgreater",\\
\quad\quad "outcome": "\textless success or failure\textgreater",\\
\quad\quad "brief\_explanation": "short explanation of the outcome"\\
\quad \},\\
\quad "lessons": [\\
\quad\quad \{\\
\quad\quad\quad "lesson\_type": "success\_pattern \textbar{} failure\_cause \textbar{} intent\_cue \textbar{} corrective\_hint",\\
\quad\quad\quad "lesson\_text": "concise reusable lesson for similar future contexts",\\
\quad\quad\quad "supporting\_evidence": ["evidence item 1", "evidence item 2"]\\
\quad\quad \}\\
\quad ],\\
\quad "edit\_proposals": [\\
\quad\quad \{\\
\quad\quad\quad "action": "add\_tentative\_edge \textbar{} reinforce\_edge \textbar{} suppress\_edge \textbar{} promote\_to\_semantic \textbar{} prune\_stale\_edge",\\
\quad\quad\quad "target": "edge, relation, or lesson-derived pattern",\\
\quad\quad\quad "update\_value": "increase confidence, decrease weight, or add tentative relation",\\
\quad\quad\quad "confidence": "low \textbar{} medium \textbar{} high",\\
\quad\quad\quad "supporting\_evidence": ["evidence item 1", "evidence item 2"],\\
\quad\quad\quad "rationale": "why this proposal may improve future recommendations"\\
\quad\quad \}\\
\quad ]\\
\}
\end{quote}
\end{tcolorbox}

\end{document}